\RequirePackage{fix-cm}
\documentclass[smallextended]{svjour3}       
\smartqed  

\usepackage[utf8]{inputenc}
\usepackage[ruled,vlined]{algorithm2e}
\usepackage{listings}
\usepackage[backend=bibtex]{biblatex}
\usepackage{subfigure}
\usepackage{siunitx}
\usepackage{mathtools}
\usepackage{amsmath}
\usepackage{amssymb}
\usepackage{dsfont}
\usepackage{graphicx}
\usepackage{adjustbox}
\usepackage{xurl}
\usepackage[dvipsnames]{xcolor}

\begin{document}

\title{Fair Interpretable Representation Learning with Correction Vectors}


\author{Mattia Cerrato \footnote{These authors contributed equally.} \and
        Alesia Vallenas Coronel \and
        Marius K\"{o}ppel $^\star$ \and
        Alexander Segner $^\star$ \and
        Roberto Esposito \and
        Stefan Kramer
}

\authorrunning{M. Cerrato et al.}
\institute{
     M. Cerrato, A. Vallenas Coronel, M. K\"oppel, A. Segner, S. Kramer \at
     Johannes Gutenberg-Universit\"{a}t Mainz \\
     Institut f\"{u}r Informatik \\
     \email{\{mcerrato,avallenas,mkoeppel,alsegner,kramerst\}@uni-mainz.de} \and
    R. Esposito \at
    Università di Torino \\
    Dipartimento di Informatica \\
    \email{roberto.esposito@unito.it} \and
}

\date{Received: date / Accepted: date}

\maketitle

\begin{abstract}
Neural network architectures have been extensively employed in the fair representation learning setting, where the objective is to learn a new representation for a given vector which is independent of sensitive information. Various representation debiasing techniques have been proposed in the literature. However, as neural networks are inherently opaque, these methods are hard to comprehend, which limits their usefulness. We propose a new framework for fair representation learning that is centered around the learning of ``correction vectors'', which have the same dimensionality as the given data vectors. Correction vectors may be computed either explicitly via architectural constraints or implicitly by training an invertible model based on Normalizing Flows. We show experimentally that several fair representation learning models constrained in such a way do not exhibit losses in ranking or classification performance. Furthermore, we demonstrate that state-of-the-art results can be achieved by the invertible model. Finally, we discuss the law standing of our methodology in light of recent legislation in the European Union.
\end{abstract}


\section{Introduction}
\label{sec:intro}
Recent breakthrough results in artificial intelligence and machine learning, in particular with deep neural networks (DNNs), 
also push their usage in tasks where the well-being of individuals is at stake, such as college admissions or loan assignment. In this situation, one needs to be concerned with the \emph{fairness} of a neural network, i.e. whether it is relying on sensitive, law-protected information such as ethnicity or gender to undertake its decisions. The usual approach here is to either constrain the output of the network \cite{deltr, fair_pair_metric} so that it is fair under some definition or to remove information about the sensitive attribute from the model's internal representations \cite{xie2017controllable, cerrato2020pairwise, mcnamara2017provably, Louizos2016TheVF}. The latter techniques might be referred to as ``fair representation learning'' \cite{zemel2013learning}. These methodologies learn a projection $f: \mathcal{X} \to \mathcal{Z}$ into a latent space where it can be shown that the information about $s$ is minimal \cite{cerrato2020constraining}. This can be evaluated experimentally to correlate with what is known as {\em group fairness} under different definitions and metrics \cite{cerrato2020pairwise, Louizos2016TheVF}. In group fairness, one is concerned with e.g. avoiding situations where a machine learning model might assign positive outcomes with different rates to individuals belonging to different groups (disparate impact \cite{zafar2017fairness}). Another problematic scenario can be observed when a model displays different error rates over different groups of people (disparate mistreatment \cite{zafar2017fairness}).

However, one issue in the area of fair representation learning is interpretability. The projection into a latent space makes it hard to investigate \emph{why} the decisions have been undertaken. This is in open contradiction with recent EU legislation, which calls for a ``right to an explanation'' for individuals which are subject to automatic decision systems (General Data Protection Regulation, Recital 71 \cite{malgieri2020gdpr}). In this context, neural networks that are \emph{fair} might still not be \emph{trustworthy} enough to fulfill increasingly strict legal requirements. 

\begin{figure}[t!]
    \centering
    \includegraphics[scale=0.5]{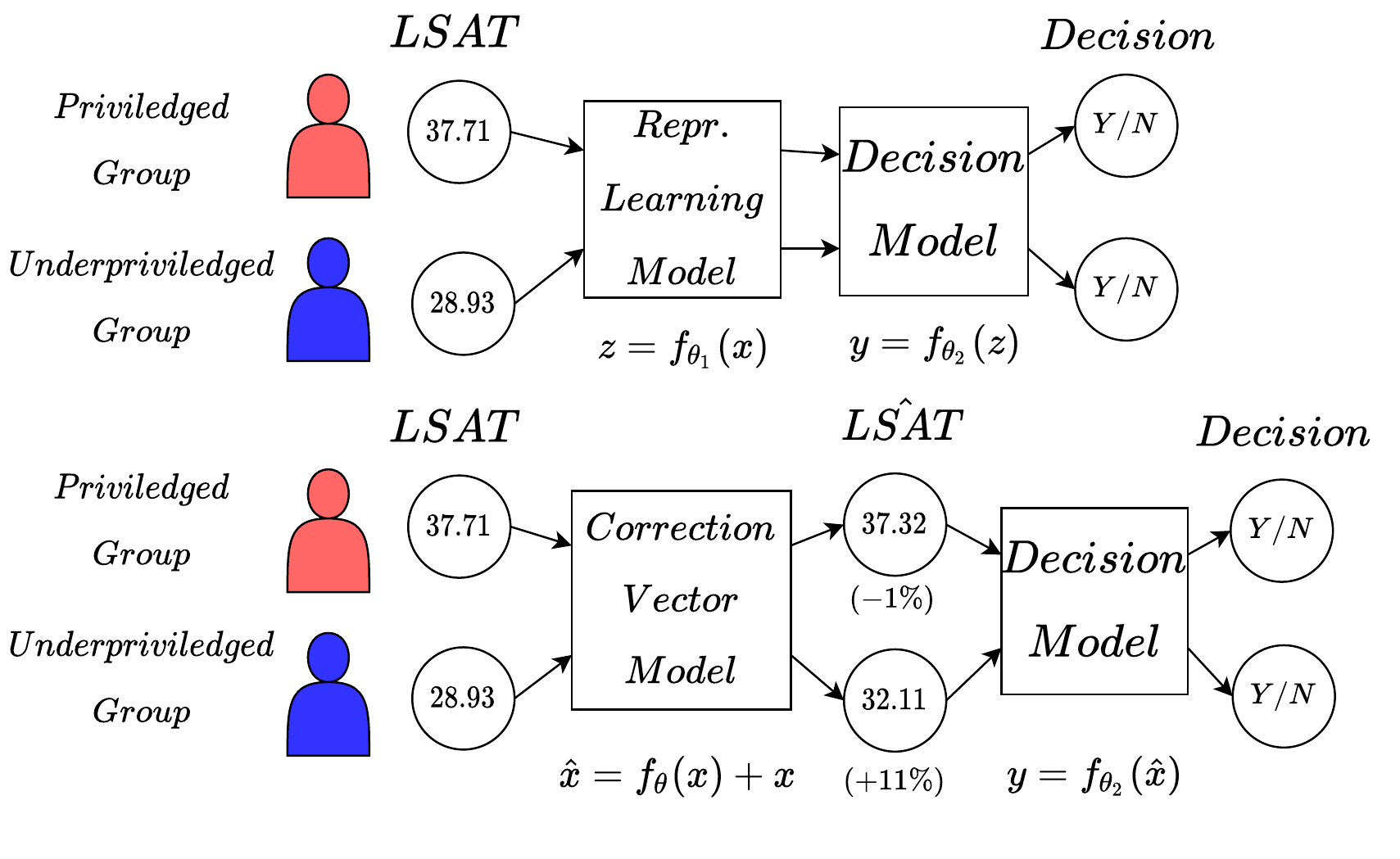}
    \caption{A regular fair representation learning algorithm (top) learns a mapping $z$ into an opaque feature space and then a decision $y$. Our correction vector framework (bottom) learns a debiased but interpretable representation $\hat{x}$. This provides individual users and analysts with further insight into the debiasing process. The correction displayed is the average correction learned by one of our models on the LSAT feature of the Law Students dataset (see Section~\ref{sec:experiments}).}
    \label{fig:example}
\end{figure}

The interpretability issues in deep neural networks may be summarized in three points \cite{interpretabilitysurvey1}:
\begin{enumerate}
    \item \emph{Complexity}. Employing models which are simply deeper and ``bigger'' has enabled performance breakthroughs in both Computer Vision (ResNets \cite{he2016deep}) and Natual Language Processing (the GPT family of models \cite{gpt3}). However, these models have incredibly high parameter counts (40 millions in ResNets, 100 billion in GPT-3). This fact, coupled with the complex non-linearities usually involved in deep networks, makes them un-readable by humans.
    \item \emph{Algorithmic opacity}. The training objectives involved are highly non-convex, and multiple solutions of similar quality may be obtained while keeping the same architecture and training data.
    \item \emph{Non-decomposability}: Complex engineering systems are usually understood in their fundamental parts and the interaction between them. In machine learning, decision trees are one model that is clearly decomposable: each node represents a decision over a feature on a well-defined criterion (e.g., based on the Gini index). While some interpretation of the layers of neural networks is possible, it is still hard to summarize the functionality of complex neural networks over a larger number of layers in a human-understandable fashion. 
\end{enumerate}

In this paper, we present a novel framework to constrain fair representation learning so that it is \emph{decomposable} and therefore human-readable. Our framework is centered around the concept of a \emph{correction vector}, i.e. a vector of features which is interpretable in feature space and represents the ``fairness correction'' each data point is subject to so that the results will be statistically fair. We assume vectorial data of not very high dimensionality, i.e., image/video data, data from signal processing, text, or high-dimensional omics data, e.g., are currently outside the scope of the method.
In general, our framework computes a function $f: \mathcal{X} \to \mathcal{X}$ and achieves decomposability by ``mapping back'' from the space of latent representations $\mathcal{Z}$ into the original feature space. Therefore, the overall fair representation learning is clearly divided into two stages: a fair preprocessing stage which computes feature corrections and a classification (or ranking) stage which optimizes for a given task. 
We introduce the computation of fair correction vectors in two different ways:

\begin{itemize}
    \item \emph{Explicit Computation}. We propose to add a small set of architectural constraints to commonly used fair representation learning models. These constraints, discussed in Section~\ref{sec:interpretable-net}, allow these architectures to explicitly compute a correction for each data point instead of a \emph{projection} into a non-interpretable latent space.
    \item \emph{Implicit Computation}. We leverage invertible neural network architectures (i.e., {\em Normalizing Flows} \cite{nice, dinh2017realnvp, normflows}) and present an algorithm which can map individuals belonging to different groups into a single one. The final result of this computation is new feature values for each individual as if they belonged to the same group. Here, a correction may still be computed as to understand what changes have been determined to be necessary to make individuals from different groups indistinguishable from one another. This methodology is discussed in Section~\ref{sec:nf}.
\end{itemize}

Our contributions can be summarized as follows: (i) We propose a new theoretical framework for fair representation learning based around the concept of \emph{correction vectors} (Section~\ref{sec:framework}). (ii) This framework may be exploited to constrain existing fair representation learning approaches so that they \emph{explicitly compute} correction vectors. We describe the relevant techniques in Section~\ref{sec:interpretable-net}. We constrain four different classification and ranking models in such a way and show that losses on fairness and relevance metrics with respect to their non-interpretable versions are negligible. (iii) We show how to \emph{implicitly compute} correction vectors by employing a pair of Normalization Flow models. This method achieves state-of-the-art results in fair classification (and ranking) and is described in Section~\ref{sec:nf}. We show our experimental results in Section~\ref{sec:experiments}. This section also presents the analysis of actual correction vectors. (iv) We discuss the law standing of the current state-of-the-art for fair representation learning, especially in light of the recent developments in EU legislation in Section~\ref{sec:legal}. 

\section{Related Work}
\label{sec:relatedworks}

\paragraph*{Fairness in Machine Learning.}\
The study of the concept of ``fairness'' in Machine Learning and Data Mining has a relatively long history, dating back to the 90s when Friedman and Nisselbaum \cite{friedman1996bias} first expressed concern about automatic decision-making performed by ``machines''. The reasoning here is that automatic decision making should still give a reasonable chance to appeal or discuss the decision. Furthermore, Friedman and Nisselbaum posit that there is a concrete risk of discrimination and ``unfairness'' which should be defined with particular attention to systemic discrimination and moral reasoning. Since then, various authors (see e.g. Verma and Rubin~\cite{verma2018fairness}; Mehrabi et al. \cite{mehrabi2021survey} for a survey) have coalesced around a definition involving protected and unprotected groups. Individuals belong to a ``protected group'' if their innate characteristics have been the subject of systemic discrimination in the past, and a ML algorithm can be said to be group-fair if e.g. it assigns positive outcomes in a balanced fashion between groups (see e.g. Zafar et al. \cite{zafar2017fairness} for a discussion of actionable group fairness definitions). For this reason, group fairness techniques have also been characterized as ``algorithmic affirmative action'' by some authors \cite{affirmativeaction, zehlike2018reducing}. Research on this topic has historically focused on pre-processing methods \cite{kamiran2009classifying}, fair classification techniques via regularization \cite{kamishima2012fairness, zafar2017fairness, dwork2012fairness} and rule-based models \cite{ruggieri2010data}. Fair ranking has also attracted more attention in recent years, in which both regularization techniques \cite{zehlike2018reducing, cerrato2020pairwise, fair_pair_metric} and post-processing \cite{zehlike2017fa} have been considered. More theoretical research has focused on the incompatibility of fairness definition and the calibration property \cite{pleiss2017calibration}, while another recent family of approaches strives to ground the group fairness concept in casual and counterfactual reasoning \cite{kusner2017counterfactual, wu2019counterfactual}.

\paragraph*{Fair Representation Learning.}\ 
The field of fair representation learning focuses on the learning of representations which are invariant to a sensitive attribute. One of the first contributions to the field by Zemel et al. \cite{zemel2013learning} employed probabilistic modelling. Representation learning techniques based on neural networks have also become popular. Various authors here have taken inspiration from a contribution by Ganin et al. \cite{ganin2016jmlr} that focuses on domain adaptation. Ganin et al. propose a ``gradient reversal layer'', which may be employed to eliminate information about a source domain when the task is to perform well on a separate but related target domain. When applying this methodology in fair machine learning, the concept of ``domains'' is adapted to be different values of a sensitive attribute. A line of research has focused on gradient reversal-based models by developing theoretical grounding \cite{xie2017controllable, mcnamara2017provably}, architectural extensions \cite{cerrato2020constraining}, and adaptations to fair ranking scenarios \cite{cerrato2020pairwise}. Other fair representation learning strategies are based on Gretton et al.'s Maximum Mean Discrepancy (MMD)  \cite{gretton2012kernel}, a kernel-based methodology to test whether two data samples have been sampled from different distributions. The relevance of this test to the fair representation learning setting is that it may be formulated in a differentiable form: as such, it has been employed in both domain adaptation \cite{tzeng2014deep} and fairness \cite{Louizos2016TheVF}.

\paragraph{Real NVP.} 
The real NVP (real Non-Volume Preserving transformations) architecture used in this work to obain fair representations from biased data is a special kind of normalizing flow, a class of learning algorithms designed to perform transformations of probability densities.
Recent work has shown that such transformations can be learned by deep neural networks, see e.g. NICE \cite{nice} and Autoregressive Flows \cite{autoregressive}. 
Furthermore it has been shown that evaluating likelihoods via the change-of-variables formula can be done efficiently by employing the real NVP architecture, which is based on invertible coupling layers \cite{dinh2017realnvp}.
In domain adaptation, the recently developed AlignFlow \cite{alignflow}, which is a latent variable generative framework that uses normalizing flows \cite{normflows,nice,dinh2017realnvp}, has been used to transform samples from one domain to another.
The data from each domain is modeled with an invertible generative model having a single latent space over all domains.
In the context of domain adaptation, the domain class is also known during testing, which is however not the case for fairness.
Nevertheless, the general idea of training two Real NVP models is used for AlignFlow as well for the model proposed in this paper.

\section{The Correction Vector Framework}
\label{sec:framework}

In this section we describe our framework for interpretable fair representation learning.
Our framework makes interpretability possible by means of computing~\emph{correction vectors}.
Commonly, the learning of fair representations is achieved by learning a new feature space $\mathcal{Z}$ starting from the input space $\mathcal{X}$.
To this end, a parameterized function $f_\theta(x)$ is trained on the data and some debiasing component is included that takes care of the sensitive data $s$.
After training, debiased data is available by simply applying the learned function $z = f_\theta(x)$.
Any off-the-shelf model can then be employed on the debiased vectors.
Various authors have investigated techniques based on different base algorithms.

The issue with the aforementioned strategy is one of interpretability.
While it is possible to guarantee \emph{invariance to the sensitive attribute} -- with much effort -- by showing that classifiers trained on the debiased data would not be able to predict the sensitive attribute, there is still no interpretation for each dimension in the latent representation $\mathcal{Z}$.
Depending on the relevant legislation, this can severely limit the applicability of fair representation learning techniques in industry and society.
Our proposal is to mitigate this issue by instead \emph{learning fair corrections} for each of the dimensions in $\mathcal{X}$.
Fair corrections are then added to the original vectors so that the semantics of the algorithm are as clear as possible.
For each feature, one can obtain a clear understanding of how that feature has been changed as to counteract the bias in the data.
Thus, we propose to learn the latent feature space $\mathcal{Z}$ by learning fair corrections $w$: $w = f_\theta(x)$ and $z = w + x$. 

In our framework, correction vectors may be learned with two different methodologies. Explicit computation requires constraining a neural network architecture so that it does not leave feature space and is presented in Section~\ref{sec:interpretable-net}. Different neural fairness methodologies may be constrained in such a way. Implicit computation relies on a pair of invertible normalizing flow models to map individuals belonging into different groups into a single one. Here a correction vector may still be computed as the new representations are still interpretable in feature space. This methodology is explained in Section~\ref{sec:nf}.

\subsection{Explicit Computation of Correction Vectors with Feedforward Networks}
\label{sec:interpretable-net}

\begin{figure}
    \centering
    \includegraphics[scale=0.65]{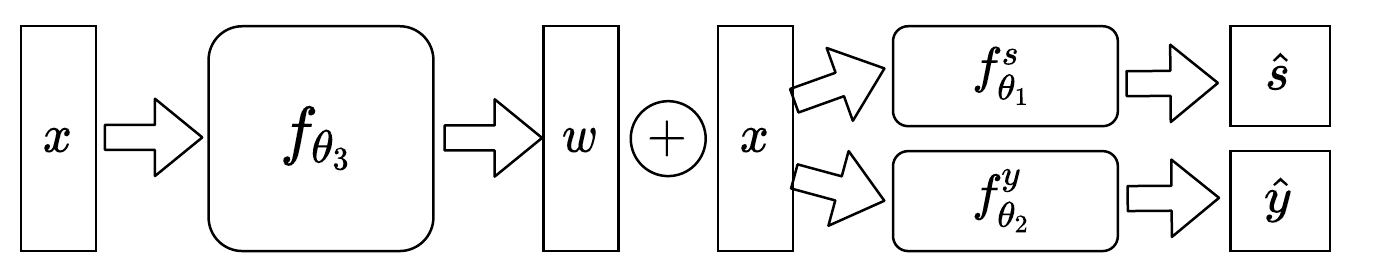}
    \caption{A gradient reversal-based neural network constrained for interpretability to be part of the proposed framework. The vector $w$ matches in size with $x$, and can then be summed with the original representation $x$ and analyzed for interpretability. This architectural constraint can be applied to other neural architectures.}
    \label{fig:architecture}
\end{figure}

It is very practical to modify existing neural network architectures so that they can belong in the aforementioned framework.
While there are some architectural constraints that have to be enforced, the learning objectives and training algorithms may be left unchanged.
The main restriction is that only ``autoencoder-shaped'' architectures may belong in our framework.
Plainly put, the depth of the network is still a free parameter, just as the number of neurons in each hidden layer. However, to make interpretability possible, the last layer in the network must have the same number of neurons as there are features in the dataset.
In an auto-encoding architecture, this makes it possible to train the network with a reconstruction loss that aims for the minimization of the difference between the original input $x$ and the output $\hat{x} = f(x)$, where $f$ is a neural network model. However, our framework only introduces the aforementioned architectural constraint and is not restricted to a specific training objective.
On top of this restriction, we also add a parameter-less ``sum layer'' which adds the output of the network to its input, the original features.
Another way to think about the required architecture under our framework is as a skip-connection in the fashion of ResNets \cite{he2016deep} between the input and the reconstruction layer (see Figure~\ref{fig:architecture}).

Constraining the architecture in the aforementioned way has the effect of making it possible to interpret the neural activations of the last layer in feature space. 
As mentioned above, our framework is flexible in the sense that many representation learning algorithms can be constrained so to enjoy interpretability properties.
To provide a running example, we start from the debiasing models based on the Gradient Reversal Layer~\cite{ganin2016jmlr} originally introduced in the domain adaptation context and then employed in fairness by various authors \cite{ mcnamara2017provably, xie2017controllable}.
The debiasing effect here is enforced by training a subnetwork $f^s_{\theta_1}(z)$ to predict the sensitive attribute.
Another sub-network learns to predict $\hat{y} = f^y_{\theta_2}(z)$.
Both networks are connected to a main ``feature extractor'' $z = f_{\theta_3}(x)$.
The two models are pitted against one another in extracting useful information for e.g. classification purposes (estimating $p(y\mid\hat{x})$) and removing information about $s$ (which can be understood as minimizing $I(\hat{x}, s)$, see \cite{cerrato2020constraining}) by inverting the gradient of $f^s_{\theta_1}(z)$ when backpropagating through the main feature extractor network $f_{\theta_3}(x)$.
Here no modification is needed to the learning algorithm, while the architecture has to be restricted so that the length of the $\hat{x}$ vector is the same as the original features $x$.
One concerning factor is whether the neural activations can really be interpreted in feature space, as features can take arbitrary values or be non-continuous (e.g., categorical).
We circumvent this issue by coupling the commonly employed feature normalization step and the activation functions of the last neural layer.
More specifically, the two functions must map to two coherent intervals of values.
As an example, employing standard scaling (feature mean is normalized to 0, standard deviation is normalized to 1) will require an hyperbolic tangent activation function in the last layer.
The model will then be enabled in learning a negative or positive correction depending on the sign of the neural activation.
It is still possible to use sigmoid activations when the features are normalized to $[0, 1]$ by means of a min-max normalization (lowest value for the feature is 0 and highest is 1).
Summing up, the debiasing architecture by Ganin et al. can be constrained for explicit computation of correction vectors via the following steps:

\begin{enumerate}
    \item Normalize the original input features $x_{raw}$ via some normalization function $x = g(x_{raw})$.
    \item Set up the neural architecture so that the length of $w = f_{\theta_3}(x)$ is equal to the length of $x$.
    \item Add a skip-connection between the input and the reconstruction layer.
\end{enumerate}

After training, the corrected vectors $z = f_{\theta_3}(x) + x$ and the correction vectors $w = f_{\theta_3}(x)$ can be interpreted in feature space by computing the inverse normalization $\hat{x}_{raw} = g^{-1}(z)$ and $w_{raw} = g^{-1}(w)$.  

Other neural algorithms can be modified similarly so to belong in the interpretable fair framework, and similar steps can be applied to, e.g., the Variational Fair Autoencoder  \cite{Louizos2016TheVF} and the variational bound-based objective of a related approach  \cite{moyer2018invariant}. As previously mentioned, our framework does not require a specific training objective and is therefore flexible. We show this by focusing our experimental validation on extending four different fair representation learning approaches so that they may compute correction vectors. Our experimentation on a state-of-the-art fair ranking model (AdvDR \cite{cerrato2020pairwise}), a fair classifier (AdvCls~\cite{xie2017controllable, cerrato2020constraining, mcnamara2017provably}), the aforementioned Variational Fair Autoencoder (VFAE~\cite{Louizos2016TheVF}) and a listwise ranker (DELTR~\cite{deltr}).
\subsection{Implicit Computation of Correction Vectors with Normalizing Flows}
\label{sec:nf}

\begin{figure}[!ht]
    \centering
    \includegraphics[width=0.9\textwidth]{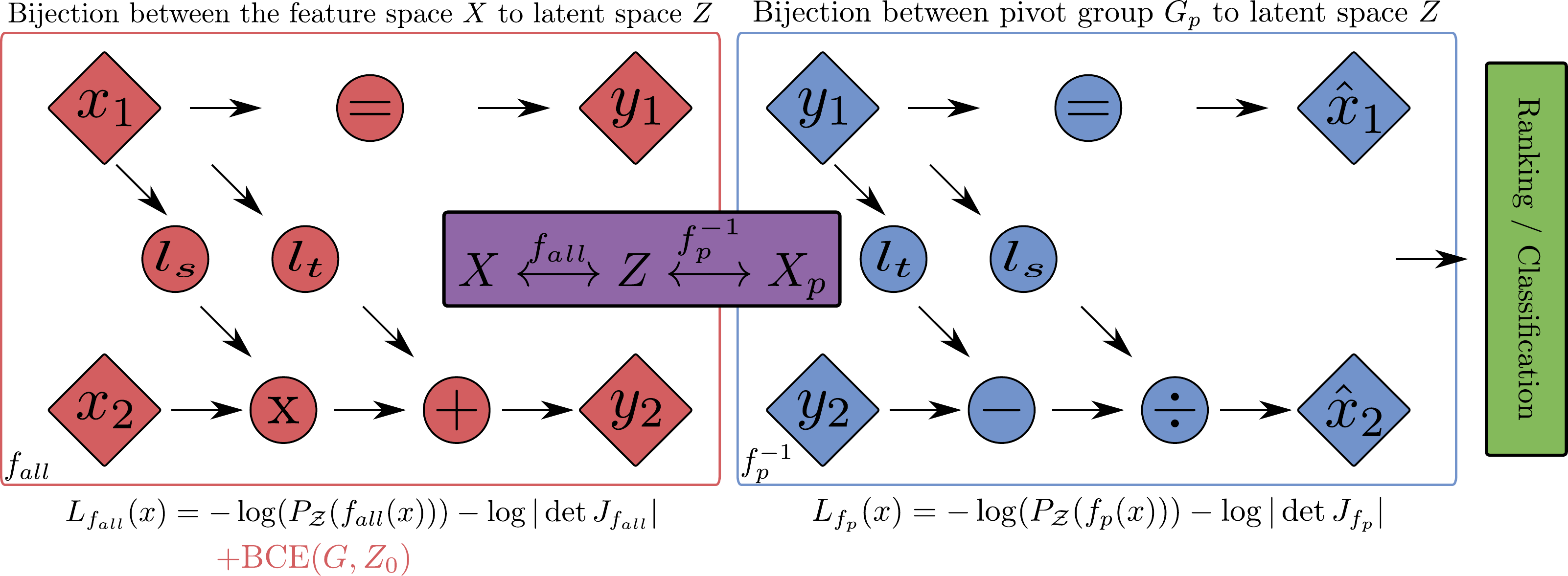}
    \caption{Sketch of the FairNF model. The two loss functions applied to each Real NVP network are indicated with $L_{f_{all}}$ and $L_{f_{p}}$. For the FairNF+BCE model an additional loss term (in this case binary cross entropy $\text{BCE}(G,Z_0)$) will be applied to the first Real NVP network (red box). This loss is employed to predict the sensitive attribute from the first feature of the latent space $Z_0$. Setting this component to a constant while evaluating the whole model, including the ranking / classification part, breaks the bijection chain and further improves the fairness of the model. Pivot group $G_p$ may be chosen to be any group in the dataset. Here $l_t$ and $l_s$ are the translation and scaling layers of the two networks.}
    \label{fig:nvp_model}
\end{figure}

This model builds upon the real NVP (real-valued Non-Volume Preserving transformations) architecture \cite{dinh2017realnvp}. For a brief introduction to this method, see Section \ref{sec:nvp_general} in the supplementary material.

Our proposed method makes use of this methodology to create fair representations that make it hard to distinguish different sensitivity groups $G_1 \dots G_{\mid s \mid}$ which are identified by their corresponding value of the sensitive attribute $s$. The learned transformation maps back into feature space and can therefore be employed to \emph{implicitly} compute correction vectors.

In the following, let $\mathcal X$ be the feature space of a dataset $\mathfrak D$ biased towards the sensitive attribute $s$ that can take $\mid s \mid$ different values. We assume that the dataset is drawn from an unknown distribution function $P_{\mathcal X}:\mathcal X\to[0,\infty)$ with each sensitivity group $G_i\subset\mathfrak D$ being drawn from their respective distribution function $P_{\mathcal X_{G_i}}$ on $\mathcal X_{G_i}\subseteq\mathcal X$.

In order to obfuscate the information on the sensitive attribute, we want to perform a transformation $f:\mathcal X\to\mathcal X_{G_p}$,  where $\mathcal G_p$ is any of the sensitivity groups represented in the data. Since the distributions in question are not generally known, but only sampled with the given dataset, we split the transformation into two transformations $f_{all}:\mathcal X\to\mathcal Z$ and $f_p:\mathcal X_{G_p} \to\mathcal Z$, where $\mathcal Z$ is an intermediate latent space on which the transformed data are distributed according to a Gaussian distribution
\begin{align}
 P_{\mathcal Z}(z)=\frac1{\sqrt{2\pi}}\exp\left(-\frac12\|z\|_2^2\right).
\end{align}
This is simple to train with the loss function
\begin{align}
 L_{F}(x)=-\log(P_{\mathcal Z}(F(x)))-\log|\det J_F|
\end{align}
with $F\in\{f_{all},f_p\}$ and $J_F$ being the Jacobian of $F$. The target transformation is then constructed as
\begin{align}
 f=f_p^{-1}\circ f_{all}
\end{align}
as depicted in Figure \ref{fig:nvp_model}.

The fair representation constructed this way, however, might not be useful for a given task and does not take advantage of ground truth information which may be available. In the case that $\mathfrak D$ is a classification or ranking dataset, we add another classification or ranking model into the chain that takes the fair representation $f(x)$ for $x\in\mathcal X$ as input and predicts a target value or ranking based on these transformed data. In order to achieve good performance on both the fairness and the prediction objectives, we train the whole chain at once with a loss function $L_y$ for the prediction objective. This is done by evaluating the gradients of $L_{f_{all}}$, $L_{f_p}$, and $L_y$ separately, where $L_{f_{all}}$ is computed using only $f_{all}$, $L_{f_p}$ is evaluated using only $f_p$, and $L_y$ is evaluated using the complete chain. The parameters of the model implementing $f_{all}$ are then updated with the gradients of
\[
 L_{all}=\gamma L_{f_{all}}+L_y,
\]
those for $f_p$ are updated with the gradients of
\[
 L_p=\gamma L_{f_p}+L_y,
\]
and those of the classification or ranking model are only updated with the gradients of $L_y$, where $\gamma$ is a tuning parameter for the tradeoff between prediction and fairness.

This procedure, however, has a caveat; namely that since $f$ is diffeomorphic, information on $s$ is technically not lost, but only obscured. We find in our experiments, discussed in Section \ref{sec:experiments}, that different values of $s$ are reasonably hard to distinguish in a representation learned in this fashion. Nonetheless, a further modification can improve the fairness still.
To this end, we introduce a projection function $f_{\text{pr}}:\mathcal Z\to\mathcal Z$ which sets the first component of $z\in\mathcal Z$ to a constant, e.g. $f_{\text{pr}}((z_1,z_2,...,z_D)^T)=(0,z_2,...,z_D)^T$. This reduces the degrees of freedom of the modified transformation
\begin{align}
 \tilde f=f_p^{-1}\circ f_{\text{pr}}\circ f_{all},
\end{align}
which is not invertible. The goal is now to maximize the overlap of $s$ with the lost degree of freedom to make sure that as little information on $s$ is present in the fair representation as possible. This is done by adding another loss term to $L_{f_{all}}$ that is aimed at predicting the sensitive attribute in the removed dimension in $\mathcal Z$. 

This leads to three different models, the first one being the base model, which only consists of the normalizing flow followed by a prediction network, which in the following we call FairNF. The other two models are the FairNF with the projection on $\mathcal Z$ included (FairNF+$f_{\text{pr}}$), and then the FairNF+$f_{\text{pr}}$ with a binary cross entropy (BCE) loss for $L_y$, called FairNF+BCE\footnote{In general, there is no restriction on the loss, but for convenience we consider BCE in the following.}. If the predictive model is a ranker, we call the complete model FairNF and in the case of classification we call it FairNFCls. 

Finally, if the feature space $\mathcal X$ is a vector space (as is usually the case), these three models can be used to compute correction vectors. That is, since the fair representation lives in a linear subspace of $\mathcal X$, a correction vector $w$ for any vector $x\in\mathcal X$ is given by
\begin{align}
    w=f(x)-x.
\end{align}

Therefore, these models do not explicitly compute corrections but are still mapping back into the original feature space $\mathcal X$. We call this technique \emph{implicit} computation of correction vectors. 

After learning the latent representation $\mathcal Z$, it is possible to chain either a classifier or a ranker depending on the data at hand. The model may then be trained in an end-to-end fashion. In the classification experiments reported in Section~\ref{sec:experiments}, we employed a simple neural classifier trained via cross-entropy. When dealing with ranking data, we instead employed a pairwise strategy inspired by the DirectRanker approach \cite{koppel2019pairwise}. This model is able to learn a total quasiorder in feature space and is competitive with more complex listwise approaches both in performance and in fairness \cite{cerrato2020pairwise}. Our chained ranking model is therefore defined as follows:

\begin{align}
    f(x_1, x_2) = \sigma(g(x_1), g(x_2)),
\end{align}

where $g$ is a fully-connected neural network and $\sigma$ is a antisymmetric, sign-conserving function such as the hyperbolic tangent. The ranking loss is then simply the squared error with respect to the indicator function $\Delta y = \mathcal{I}_{y_1 \geq y_2}$:

\begin{align}
    L_{y} = (\Delta y - f(x_1, x_2))^2
\end{align}

To the end of better understanding which FairNF architecture behaves the best with respect to fairness and invariant representations, we have performed experimentation on a simple synthetic dataset. For space and presentation reasons, we include our findings in the supplementary material (Section~\ref{sec:knn_mixture}).

\section{Experiments}\label{sec:experiments}

For evaluating the different models, we performed experiments on ranking and classification datasets commonly used in the fairness literature.
We evaluated each model using different relevance and fairness metrics and evaluated everything on a 3 times 3 fold gridsearch to find the best hyperparameter setting.
In the following we describe our experimentation in detail, including the employed evaluation metrics and datasets. 

\subsection{Evaluation Metrics}\label{sec:metric}
Besides the well-known AUROC, which we will call AUC in the following, different fairness and performance measures are used to evaluate the different algorithms.
In the following, we will give a short overview of the metrics used in this work.

\subsubsection{Ranking Metrics}

\paragraph*{The NDCG Metric.}\label{sec:ndcg}\ The normalized discounted cumulative gain of top-$k$ documents retrieved (or NDCG@$k$) is a commonly used measure for performance in the field of learning to rank.
Based on the cumulative gain of top-$k$ documents (DCG@$k$), the NDCG@$k$ can be computed by dividing the DCG@$k$ by the ideal (maximum) discounted cumulative gain of top-$k$ documents retrieved (IDCG@$k$):
\begin{equation}
\text{NDCG@}k = \frac{\text{DCG@}k}{\text{IDCG@}k} = \frac{\sum_{i=1}^{k} \frac{2^{r(d_i)} - 1}{\log_2(i + 1)}}{\text{IDCG@}k}
\,,\notag
\end{equation}
where $d_1, d_2, ..., d_n$ is the list of documents sorted by the model with respect to a single query and $r(d_i)$ is the relevance label of document $d_i$.

\paragraph*{rND.}\label{sec:rnd}\ To the end of measuring fairness in our models, we employ the rND metric~\cite{ke2017measuring}. 
This metric is used to measure group fairness and is defined as follows: 

\begin{equation}
\text{rND} = \frac{1}{Z} \sum_{i \in \{10, 20, ...\}}^N \frac{1}{\log_{2}(i)} \left| \frac{ \mid S^{+}_{1...i} \mid}{i} - \frac{\mid S^+ \mid}{N} \right |.
\end{equation}

The goal of this metric is to measure the difference between the ratio of the protected group in the top-$i$ documents and in the overall population.
The maximum value of this metric is given by $Z$,  which is also used as normalization factor.
This value is computed by evaluating the metric with a dummy list, where the protected group is placed at the end of the list. This biased ordering represents the situation of ``maximal discrimination''. 

This metric also penalizes if protected individuals at the top of the list are over-represented compared to their overall representation in the population.

\paragraph*{Group-dependent Pairwise Accuracy.}\label{sec:gpa}\ Let $G_1, ..., G_K$ be a set of $K$ protected groups such that every instance inside the dataset $\mathfrak D$ belongs to one of these groups. The \emph{group-dependent pairwise accuracy} \cite{fair_pair_metric} $A_{G_i > G_j}$ is then defined as the accuracy of a ranker on instances which are labeled more relevant belonging to group $G_i$ and instances labeled less relevant belonging to group $G_j$. Since a fair ranker should not discriminate against protected groups, the difference $|A_{G_i > G_j} - A_{G_j > G_i}|$ should be close to zero. In the following, we call the Group-dependent Pairwise Accuracy {\em GPA}. We also note that this metric may be employed in classification experiments by considering a classifier's accuracy when computing $A_{G_i}$ and $A_{G_j}$. 

\subsubsection{Classification Metrics}

\paragraph*{Area Under Discrimination Curve.}\label{sec:AUDC}\ We take the discrimination as a measure of the bias with respect to the sensitive feature $s$ in the classification \cite{zemel2013learning}, which is given by:
\begin{equation}
\text{yDiscrim} = \left | \frac{\sum^n_{n:s_n=1} \hat{y}_n}{\sum^n_{n:s_n=1} 1} - \frac{\sum^n_{n:s_n=0} \hat{y}_n}{\sum^n_{n:s_n=0} 1} \right |, 
\end{equation}
where $n:s_n=1$ denotes that the $n$-th example has a value of $s$ equal to 1. This measure can be seen as a statistical parity which measures the difference of the proportion of the two different groups for a positive classification.
Similar to the AUC we can evaluate this measure for different classification thresholds and calculate the area under this curve.
Using different thresholds, dependencies on the yDiscrim measure can be taken into account.
We call this metric in the following AUDC (Area Under the Discrimination Curve). 
One issue with this metric is that it may hide high discrimination values on certain thresholds as they will be ``averaged away''. We show plots analyzing the accuracy/discrimination of our models tradeoff at various thresholds in Figure~\ref{fig:compare_auc} in the supplementary material. These were obtained by computing the discrimination and accuracy of our models at 20 different thresholds on the interval $[0.05, 1)$.  

\paragraph*{Absolute Distance to Random Guess.}\label{sec:ADRG}\ To evaluate the invariance of the representation with respect to the sensitive attribute, we report classifier accuracy as the absolute distance from a random guess (the majority class ratio in the dataset), which we call ADRG in the following.

\subsection{Datasets} \label{sec:datasets}
We focused our evaluation on four real-world datasets commonly employed in the fairness literature.
The COMPAS dataset was released as part of an investigative journalism effort in tackling automated discrimination \cite{machine_bias}.
The dataset contains two ground truth values.
First a risk score (from 0 to 10) of how likely a person will commit a crime and second whether that person committed a crime in the future.
For the ranking experiments, we took the risk score as the target value, while for the classification ones, we tried to predict if the person committed a crime in the future.
In terms of correction vectors analysis (see Section \ref{sec:cor_ana}) we evaluated the feature ``prior crimes'' and how much it was changed by the interpretable models to make the representations fair.

Moreover, the Adult dataset is used, where the ground truth represents whether an individual's annual salary is over 50K\$ per year or not \cite{adult}. 
It is commonly used in fair classification, since it is biased against gender \cite{Louizos2016TheVF,zemel2013learning,cerrato2020constraining}.
For the correction vectors analysis, the feature ``capital gain'' was analyzed.

The third dataset used in our experiments is the Bank Marketing Data Set \cite{banks}, where the classification goal is whether a client will subscribe a term deposit.
The dataset is biased against people under 25 or over 65 years.
The ``employment variation rate'' was studied for the correction vectors analysis.

The last dataset we used is the Law Students dataset, which contains information relating to 21,792 US-based, first-year law students and was collected to the end of understanding whether the Law Students Admission Test in the US is biased against ethnic minorities~\cite{law_student}. As done previously~\cite{zehlike2018reducing}, we subsampled 10\% of the total samples while maintaining the distribution of gender and ethnicity, respectively. We used ethnicity as the sensitive attribute, while the ground truth of the dataset is to sort students based on their predicted academic performance.
Here, we analysed the change in LSAT score during our correction vectors analysis.

\subsection{Experimental Setup}
To overcome statistical fluctuations during the experiments, we split the datasets into 3 internal and 3 external folds.
On the 3 internal folds, a Bayesian optimization technique, maximizing the fairness measure (1-rND for the ranking datasets and 1-AUDC for the classification datasets) is used to find the best hyperparameter setting.
The best setting is then evaluated on the 3 external folds. We relied on the Weights \& Biases platform for an implementation of Bayesian optimization and overall experiment tracking \cite{wandb}. 

We experiment with a state-of-the-art algorithm called Fair Adversarial DirectRanker (AdvDR in the rest of this paper), which showed good results on commonly used fairness datasets \cite{cerrato2020pairwise}, a Debiasing Classifier (AdvCls) based on gradient reversal \cite{ganin2016jmlr, cerrato2020constraining, xie2017controllable, mcnamara2017provably}, a fair listwise ranker (DELTR \cite{deltr}) and the Variational Fair Autoencoder \cite{Louizos2016TheVF}\footnote{Code was taken from \url{https://github.com/yevgeni-integrate-ai/VFAE} and included into our experimental framework.} (VFAE). 
All these methods were also constrained for explicit computation of correction vectors. In the following we will indicate the models constrained in such a way with ``Int'' before the name of the model (IntAdvDR, IntVFAE, etc.).

Since it is from a theoretical point (to the best of our knowledge) not clear how exactly a real NVP model is treating discrete features, the implicit correction vector computation technique (FairNF for ranking and FairNFCls for classification) was trained on continuous features only\footnote{For our implementation of the models, the experimental setup, and the used datasets, see~\url{https://zenodo.org/record/5572596}}.

\subsection{Model Results}

\begin{figure*}[!ht]
\centering
\subfigure[COMPAS Ranking]{\label{fig:ranking_compas_gpa}\includegraphics[width=0.32\textwidth]{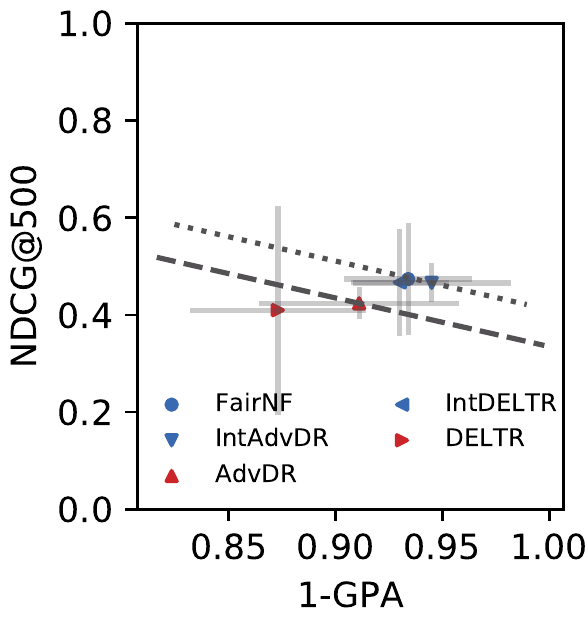}}
\subfigure[Law-Race Ranking]{\label{fig:ranking_law_gpa}\includegraphics[width=0.32\textwidth]{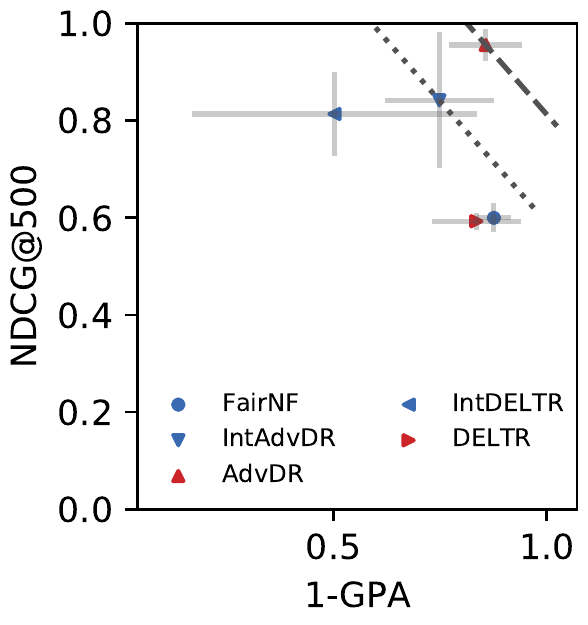}}
\subfigure[Banks Ranking]{\label{fig:ranking_banks_gpa}\includegraphics[width=0.32\textwidth]{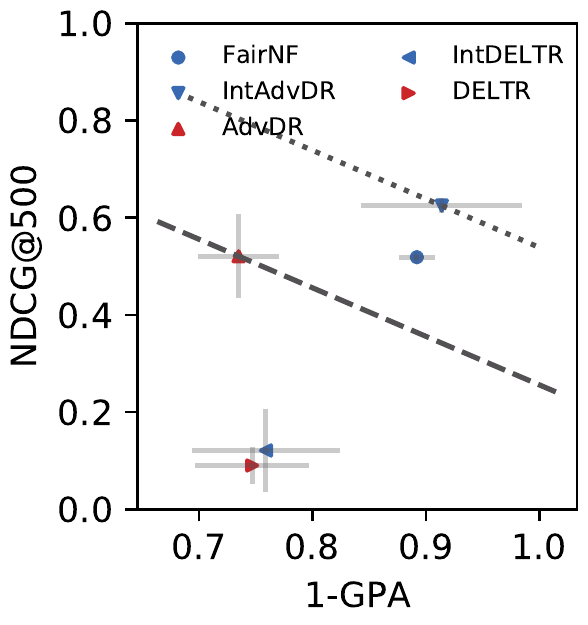}}
\subfigure[COMPAS Ranking]{\label{fig:ranking_compas_rnd}\includegraphics[width=0.32\textwidth]{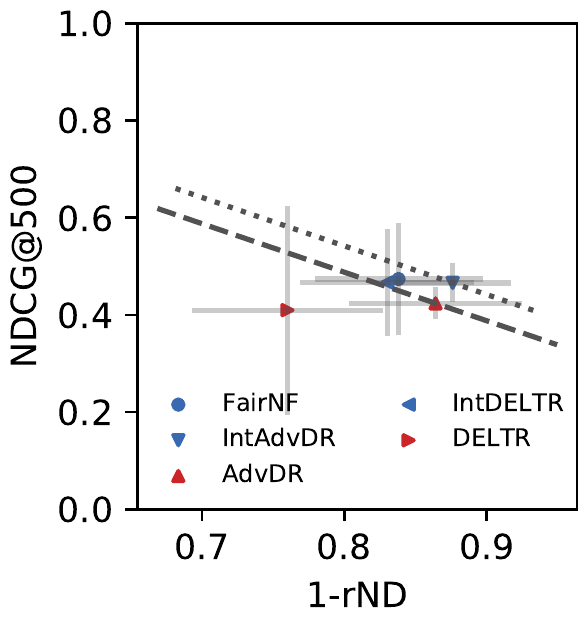}}
\subfigure[Law-Race Ranking]{\label{fig:ranking_law_rnd}\includegraphics[width=0.32\textwidth]{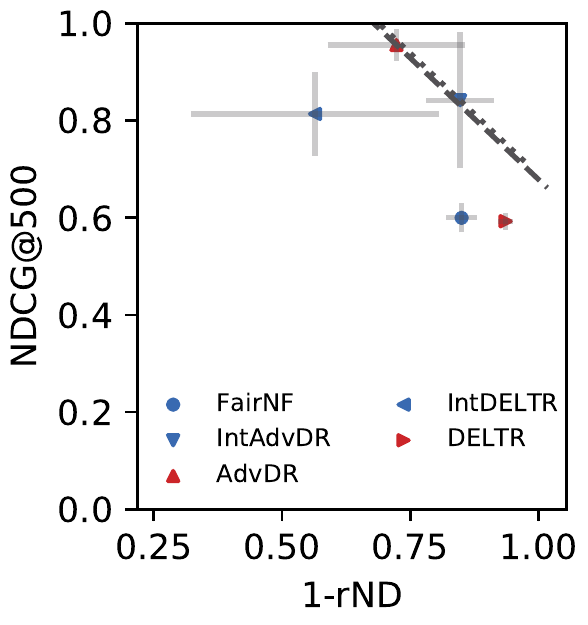}}
\subfigure[Banks Ranking]{\label{fig:ranking_banks_rnd}\includegraphics[width=0.32\textwidth]{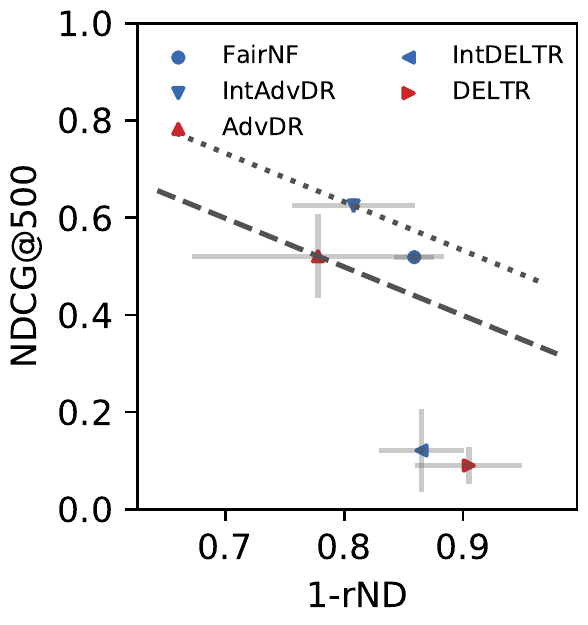}}
\caption{Results for the COMPAS, Law-Race and Banks datasets for the ranker outputs. The lines represent balanced fairness/relevance trade-offs. The dashed line represents all points with equal trade-off as the best performing comparison model, while the dotted line also takes into consideration the models we contribute. The grey lines around each model represent the error computed over 3 folds.}
\label{fig:ranking-results}
\end{figure*}

\begin{figure*}[!ht]
\centering
\subfigure[COMPAS Classification]{\label{fig:cls_compas_gpa}\includegraphics[width=0.32\textwidth]{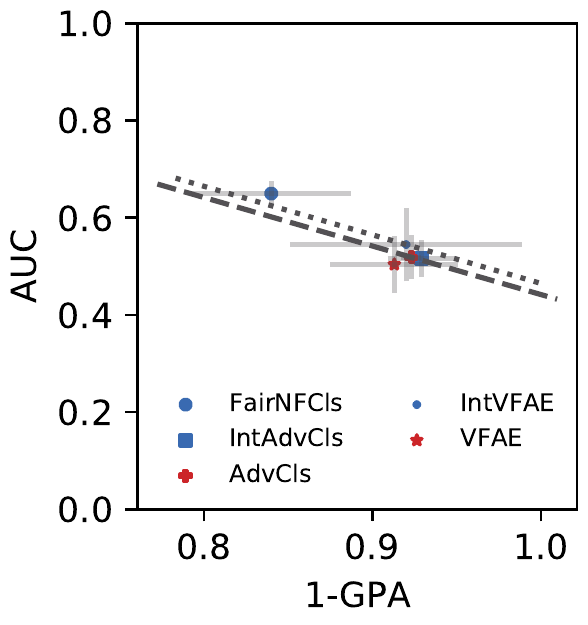}}
\subfigure[Adult Classification]{\label{fig:cls_adult_gpa}\includegraphics[width=0.32\textwidth]{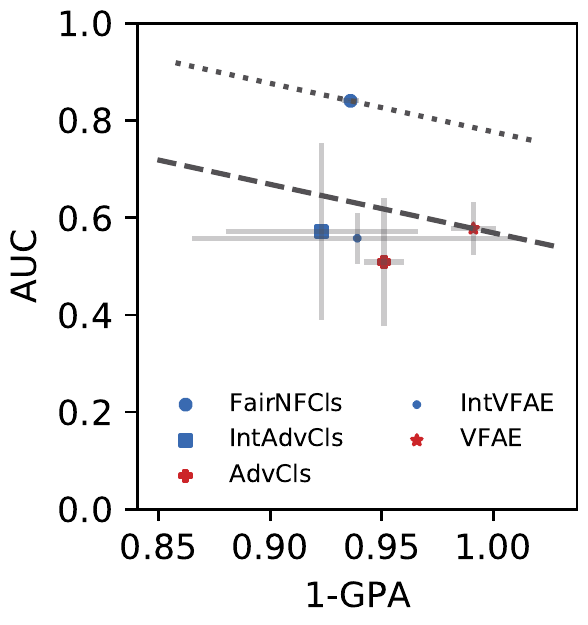}}
\subfigure[Banks Classification]{\label{fig:cls_banks_gpa}\includegraphics[width=0.32\textwidth]{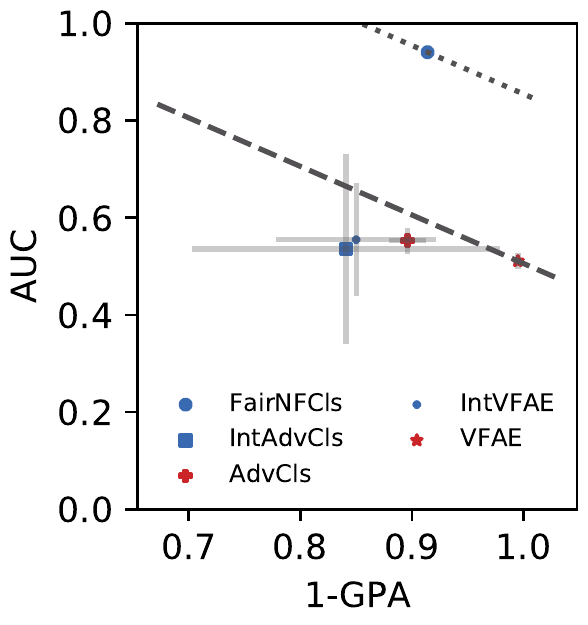}}
\subfigure[COMPAS Classification]{\label{fig:cls_compas_audc}\includegraphics[width=0.32\textwidth]{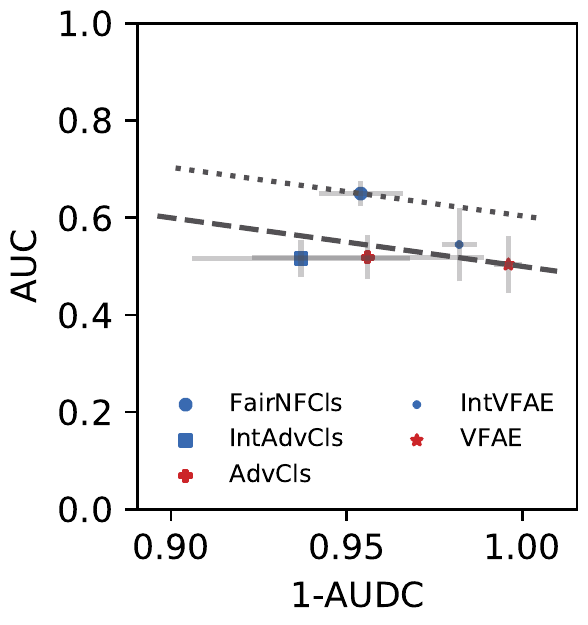}}
\subfigure[Adult Classification]{\label{fig:cls_adult_audc}\includegraphics[width=0.32\textwidth]{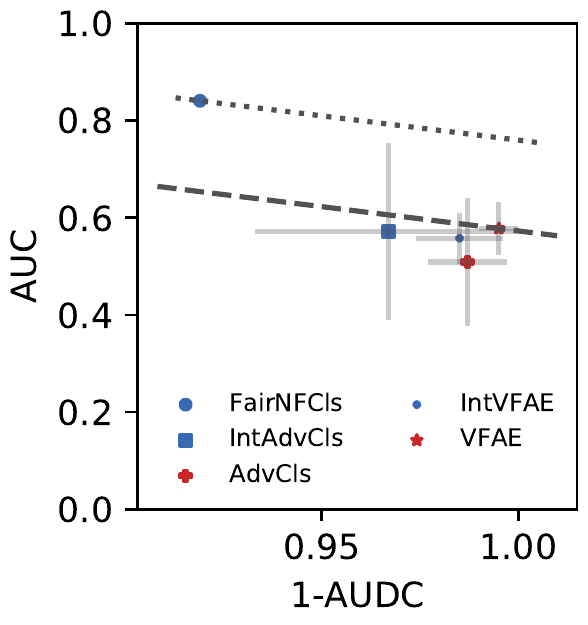}}
\subfigure[Banks Classification]{\label{fig:cls_banks_audc}\includegraphics[width=0.32\textwidth]{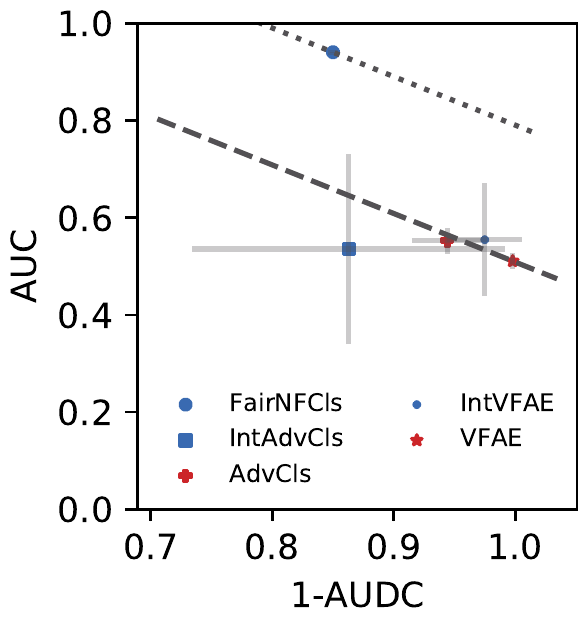}}
\caption{Results for the COMPAS, Adult and Banks datasets for the classifier outputs. The lines represent balanced fairness/accuracy trade-offs. The dashed line represents all points with equal trade-off as the best performing comparison model, while the dotted line also takes into consideration the models we contribute. The grey lines around each model represent the error computed over 3 folds.}
\label{fig:classification-results}
\end{figure*}

We report our experimentation results for fair ranking in Figure~\ref{fig:ranking-results}. Results for fair classification are included in Figure~\ref{fig:classification-results}. In all the figures we plot the line of optimal trade-off as found by a model included in the experimentation. The optimal trade-off is defined as the smallest value of $\|(1,1)-(1-m,n)\|_1$ with $(m,n)\in\{\text{(rND,nDCG),(GPA,AUC), (GPA, nDCG), (AUDC, AUC)}\}$. The line for the models already present in the literature (AdvDR, DELTR, AdvCls, VFAE) is drawn dashed; the line for the models we introduced (the explicitly constrained IntAdvDR, IntDELTR, IntAdvCls, IntVFAE; the implicit correction vector computation model FairNF/FairNFCls) is dotted. We also test the relevance/fairness tradeoff in both disparate impact (via the rND and AUDC metrics for ranking and classification respectively) and disparate mistreatment (via GPA) situations. 

Overall, we find that the proposed models can be just as good or better than the models we compare against when considering the optimal trade-off defined above. The performance of the IntAdvDR model is especially impressive, showing higher fairness and relevance than its non-interpretable counterpart in all considered scenarios but one (Figure~\ref{fig:ranking_law_gpa}). The fair classifier IntAdvCls shows slightly reduced fairness on average when compared to the AdvCls model, which is however within the margin of error. Results for DELTR and IntDELTR appear to be suffering from larger errors than the other models we considered. Our interpretable model compares favorably on COMPAS (Figures~\ref{fig:ranking_compas_gpa} and~\ref{fig:ranking_compas_rnd}) and Banks (Figures~\ref{fig:ranking_banks_gpa} and~\ref{fig:ranking_banks_rnd}) in terms of fairness; however, it had issues converging to a fair solution on the Law-Race dataset (Figures~\ref{fig:ranking_compas_rnd} and~\ref{fig:ranking_compas_gpa}). We hypothesize that this is due to overfitting. 
The implicit correction vector model displays impressive fair classification performance, finding the optimal tradeoff in all the datasets and metrics we considered (Figures~\ref{fig:cls_compas_gpa} through~\ref{fig:cls_banks_audc}, FairNFCls). It is also able to find consistently fair rankings (Figures~\ref{fig:ranking_compas_gpa} through~\ref{fig:ranking_banks_rnd}, FairNF) at the cost of slightly reduced relevance when compared to other methodologies.

\subsection{Direct Comparison of Interpretable vs. Non-Interpretable Models}\label{sec:compare_model}

\begin{figure*}[!ht]
\centering
\subfigure[1-GPA]{\label{fig:compare_gpa_ranking}\includegraphics[width=0.49\textwidth]{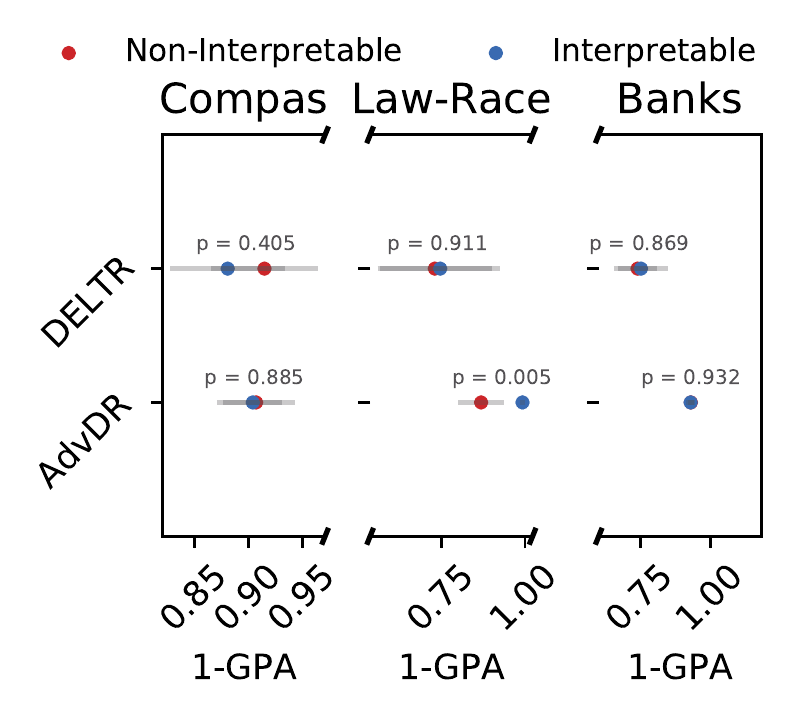}}
\subfigure[1-GPA]{\label{fig:compare_gpa}\includegraphics[width=0.49\textwidth]{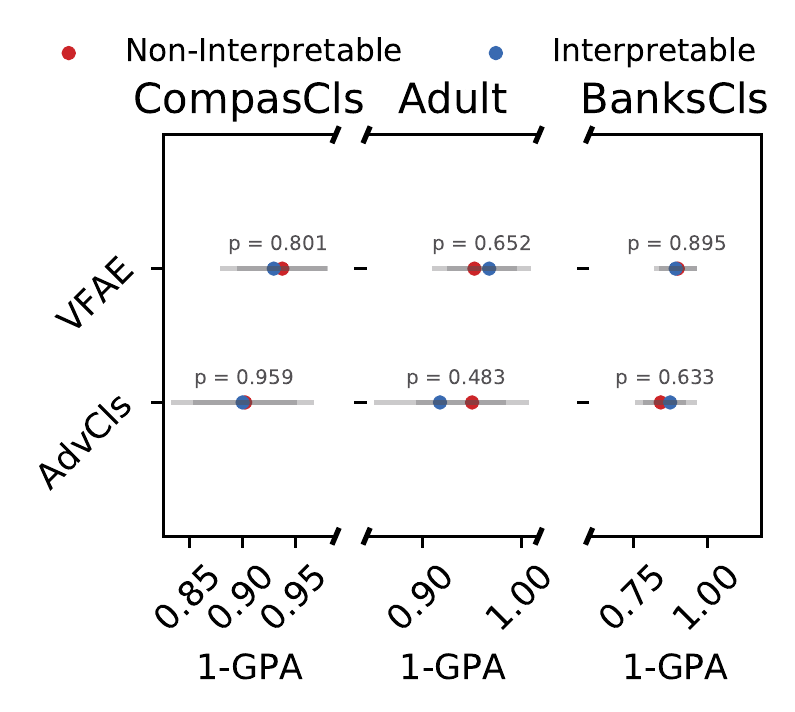}}
\subfigure[1-rND]{\label{fig:compare_rnd}\includegraphics[width=0.49\textwidth]{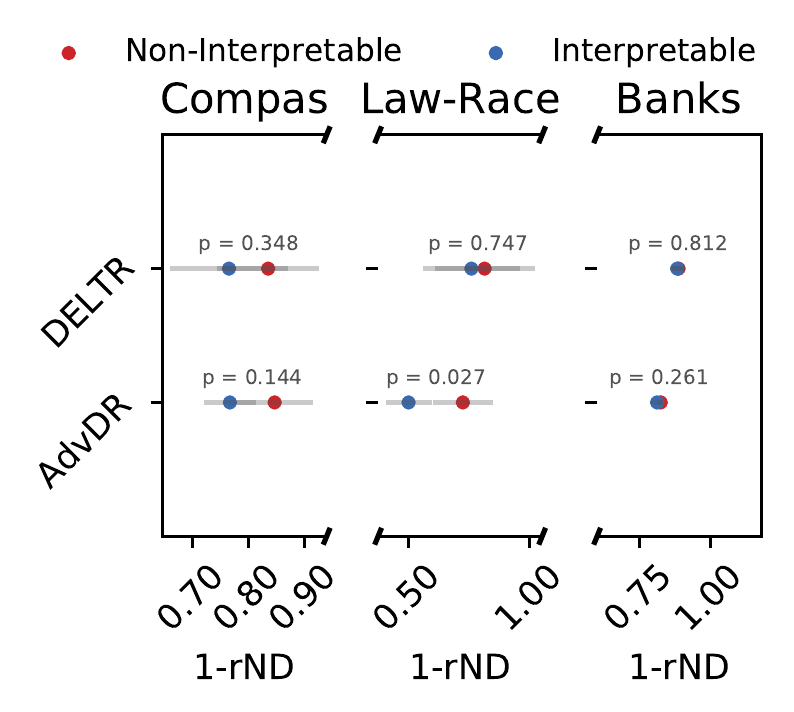}}
\subfigure[1-AUDC]{\label{fig:compare_audc}\includegraphics[width=0.49\textwidth]{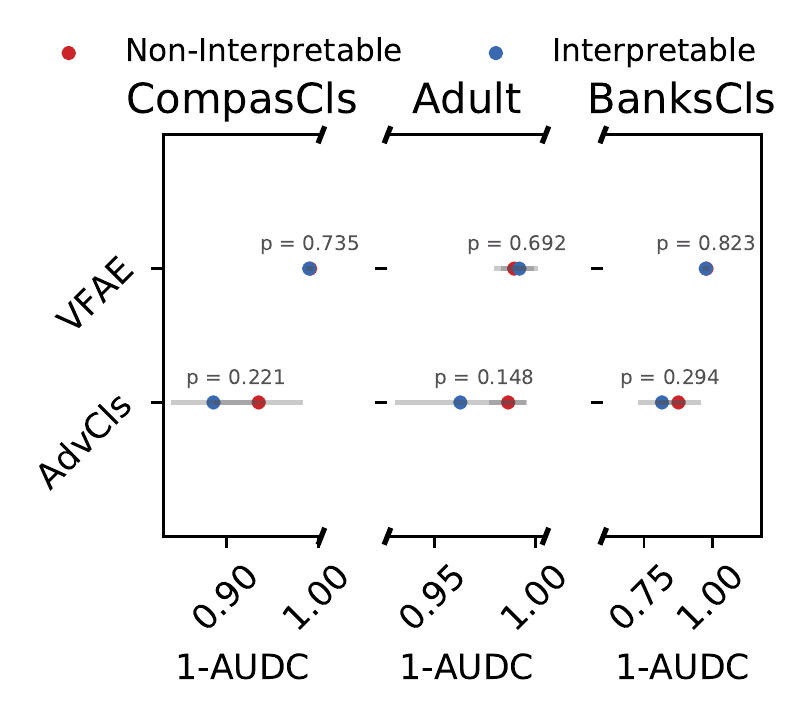}}
\caption{Results for all datasets comparing the fairness of each model with and without constraining it to be interpretable. The blue points represents all interpretable models, while the red points show the non interpretable models. The grey lines around each model represent the error computed over 3 folds.}
\label{fig:interpretable-comparison}
\end{figure*}

To the end of understanding the \emph{fairness} impact of the architectural constraints we imposed to allow for explicit computation of fairness vectors (Section~\ref{sec:interpretable-net}) we perform further experimentation. Each of the architectures that were constrained for direct computation of correction vectors were compared to their non-interpretable versions trained as described in the respective papers introducing them. We trained each interpretable/non-interpretable pair of architectures 100 times while searching for the best hyperparameters. Each dataset is split into 3 internal folds which are employed for hyperparameter selection purposes, while 15 external folds are kept separate until test time. We then analyze the average performance on the relevant fairness metrics (1-GPA, 1-rND for ranking; 1-AUDC for classification) on the external folds. We correct the variance estimation as described by Nadeau and Bengio \cite{nadeau2003inference} and perform a significance t-test. Here the number of runs averaged is the number of test (external) folds $n=15$ following the recommendations by Nadeau and Bengio, who found little benefit in higher values in terms of statistical power. We report the pairwise comparisons between interpretable and non-interpretable models in Figure~\ref{fig:interpretable-comparison}. Each comparison also reports the p-value obtained by the application of the t-test. We found that the data does not support the presence of statistically significant differences between the fairness performance of our interpretable models and the non-interpretable ones. One model/dataset combination does take exception to this, however. This exception is represented by the Law-Race dataset when learning AdvDR ranking models. In Figure~\ref{fig:compare_gpa} we observe a statistically significant advantage in GPA when employing the interpretable model, whereas this trend is reversed when considering rND (Figure~\ref{fig:compare_rnd}). We hypothesize here that our hyperparameter search, which seeks to find models with the best values of $(1-GPA) + (1-rND)$, has found an impressively fair interpretable model for 1-GPA while sacrificing rND performance. The interpretable AdvDR does not seem to suffer much in terms of rND on other datasets, which suggests that it has no issue optimizing rND in general. We conclude that our framework for explicit computation of correction vectors warrants strong consideration in situations where transparency is needed to comply with existing laws and regulations.

\subsection{Fair Representations}\label{sec:fair_rep}

In fair representation learning, the ultimate objective is removing information about the sensitive attribute in the obtained representation. We investigate this matter by training off-the-shelf rankers and classifiers on the transformed data. The task is to classify the sensitive attribute, i.e. recovering information about it. We show results for this task in Figure~\ref{fig:ranking_reps} and Figure~\ref{fig:classification_reps} of the supplementary material.
The employed models are plain Linear Regression (LR) and Random Forests (RF). We evaluate the absolute difference in accuracy from the random guess (ADRG \cite{cerrato2020constraining, cerrato2020pairwise, Louizos2016TheVF}). As explained above, this metric is defined as the absolute value of the difference between the accuracy of an external classifier and the accuracy of a classifier which always predicts the majority class. Here, the rationale is that a perfectly fair representation will force a classifier to always predict the majority class as there is no other way to obtain a higher accuracy value. Therefore, lower scores are better. We also computed the AUC of the employed classifiers. 
We observe only minimal differences between the algorithms we propose in this paper and the ones already present in the literature. One issue we observed with the VFAE algorithm is that it highly degrades external classifier performance to the point that the accuracy becomes lower than the random guess. In a binary classification setting, one could invert the decisions undertaken by the model and obtain a higher performance than random guessing: therefore, this model displays fairly high ADRG.

As an aside, gains in representation invariance with respect to the sensitive attribute are very small when considering AUC. To the best of our knowledge, this is the first time that AUC has been considered for this kind of evaluation, with previous literature focusing mostly on accuracy-based measures such as ADRG (see e.g. \cite{Louizos2016TheVF, zemel2013learning, cerrato2020constraining}). We conclude that employing AUC should be strongly considered in the future when evaluating fair representation learning algorithms.

\subsection{Correction Vector Analysis}\label{sec:cor_ana}

\begin{figure*}[!ht]
\centering
\subfigure[Direction Plot IntAdvCls]{\label{fig:dir_1}\includegraphics[width=0.49\textwidth]{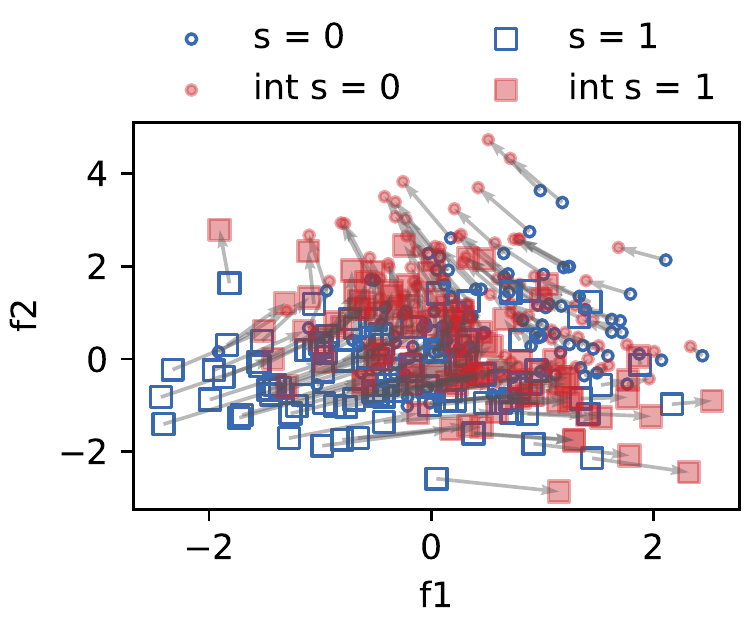}}
\subfigure[Direction Plot FairNF]{\label{fig:dir_2}\includegraphics[width=0.49\textwidth]{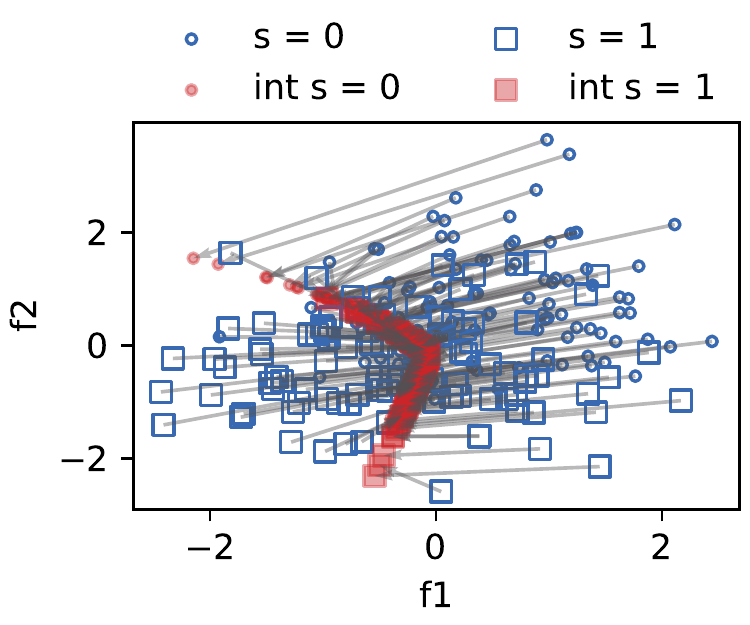}}
\caption{Direction plots for the computed corrections in a AdvCls and FairNF model trained on a two-Gaussians dataset. The two distributions have unit variance and differ only in their mean. Each distribution is akin to a different value of a binary-valued sensitive attribute, with circles representing $s=0$ and squares representing $s=1$. The original samples are plotted in \textcolor{blue}{blue}; the corrected samples are plotted in \textcolor{red}{red}. } \label{fig:direction_plots}
\end{figure*}

In this section, we offer a qualitative analysis of the obtained correction vectors for all the models introduced in the paper. To provide a more visual example, we also provide pre- and post-correction visualizations for a simple two-Gaussians dataset, which can be found in Figure~\ref{fig:direction_plots}. This dataset consists of samples from two separate Gaussian distributions which only differ in their mean. More specifically, the mean vectors are $\mu_0 = [0, 0]$ and $\mu_1 = [1, 1]$, forming two point clouds which partially overlap as we opted for an unit-valued variance for both. We employ this dataset in training an interpretable AdvCls \cite{cerrato2020constraining, xie2017controllable}, which performs explicit corrections (see Section~\ref{sec:interpretable-net}), and a FairNF model (see Section~\ref{sec:nf}). What we observe in the plots is the AdvCls model performing rather intuitive translations, which push the two Gaussian distributions closer in feature space after correction. The FairNF model instead opts to push both distributions into a line-like manifold. We reason that this is a side effect of our strategy to break the bijection feature of Normalizing Flows via discarding one of the features in the latent space $\mathcal Z$. We do note that the examples are centered around $[0, 0]$ after correction, meaning that the second Gaussian has been pushed onto the first. It is noteworthy, however, that the original distribution shapes were not kept equal, even if the variance appears to be similar.

After having built a visual intuition about how explicit and implicit corrections might look and differ, we now move on to analyzing the corrections on the real-world datasets described in Section~\ref{sec:datasets}. Our analysis was performed as follows. For each dataset, we pick a feature for which we expect correction vectors to make individuals belonging to different groups more similar. We then compute the average correction for each group and compute the average difference in the chosen feature in absolute and relative terms. A table with all average corrections for each model and dataset can be found in Figure~\ref{tab:interpretable-vects} in the Supplementary Material. 
What stands out from these results is that different models often disagree about both the sign and the intensity of the correction that should be applied. For instance, on the COMPAS dataset, we would expect a correction to reduce the difference in the average amount of previous crimes committed between white and black people. While our implicit computation methodology does just that, the VFAE constrained for explicit computation of correction vector opts to widen that gap. Nonetheless, it obtains an overall fair result (see Figures~\ref{fig:cls_compas_gpa}, \ref{fig:cls_compas_audc}, \ref{fig:compare_audc} and \ref{fig:compare_gpa}). Similar extreme corrections, which are definitely not intuitive, were computed by the interpretable AdvCls and FairNF models on the Adult dataset. One noteworthy trend is that simpler architectures computed more intuitive corrections. To be more specific, both the AdvDR and DELTR models employ only a single neural layer after computing the corrected features. In both models this layer has a single neuron which outputs the ranking scores ($p(y_1 > y_2)$ for the pairwise AdvDR and $p_{top_1}(x_i)$ for the listwise strategy DELTR). This severely constrains the complexity of the transformation learned by the models after feature vector computation and is highly in contrast with the other models, which have no such restrictions. DELTR and AdvDR are able to clearly close the gap in average feature values in all considered datasets.
The other models were also able to output reasonable corrections. For instance, our FairNF model opted to almost nullify the gap in prior crimes committed on the COMPAS dataset. However, these models appear to be less consistent in this regard -- FairNF instead opted to amplify the average difference in employment variation rate on the Adult dataset. Interestingly, this model is the strongest tradeoff performer on that same dataset (Figure~\ref{fig:cls_adult_gpa}). As previously discussed, we hypothesize that the absence of restrictions about how many layers may be included after the computation of the correction vectors may limit the intuitiveness of some of the corrections. This issue could be addressed in the future by further constraining the maximum computable value for correction vectors.
These matters notwithstanding, the opportunity to peek inside the ``black box'' of fair representation learning might also guide how a practitioner would select models: a model which computes corrections that are deemed counter-intuitive or excessive may be discarded during model selection. Therefore, our framework adds a new layer of opportunities in developing models that can be trustworthy and transparent on top of fair. 

\section{Correction Vectors: a Legal Perspective}
\label{sec:legal}

Our proposal has to be discussed in the context of the GDPR  \cite{GDPR} and the more recent regulation of AI in the EU \cite{aiact}. As an expression of Article 8 EU of the Charter of Fundamental Rights \cite{fundrights} and Article 16 of the Treaty on the Functioning of the EU \cite{eutreaty}, the GDPR protects the right to informational self-determination, according to which everyone should be able to decide which personal data one wishes to disclose and by whom it may be used. Article 22 (1) of the GDPR stipulates a basic right for data subjects “not to be subject to a decision based solely on automated processing”. Paragraph 2 entails exceptions of the general prohibition of automated decision-making. These exceptions are linked to some of the key transparency requirements of the GDPR that are outlined in Articles 13-15 GDPR  and constitute a data subject’s right to be informed. Following these transparency requirements, data subjects are to be provided with ``meaningful information about the logic involved''\footnote{Recital 63 of the GDPR; Articles 13 (2)(f), 14 (2)(g), 15 (1)(h) of the GDPR \cite{GDPR}.}.

With this legal background, two potential problems arise in the context of our proposal: Firstly, which content requirements are to be met when having to disclose “the logic involved” and, secondly, their realisation in practice\footnote{See Hoeren and Niehoff, 2018 \cite{hoeren2018ki} 1 RW 47, 55.}. In general, only the principles\footnote{The French version of the GDPR speaks explicitly of revealing only the logic that forms the basis: "la loquique qui sous-tend leur éventuel traitement automatisé". This is also the case with the Dutch version: "welke logica er ten grondslag ligt".} behind a decision have to be stated under the GDPR, not the algorithm formula per se\footnote{See Recital 63 of the GDPR \cite{GDPR}; BGHZ 200, 28; Hoeren and Niehoff, 2018 \cite{hoeren2018ki} 1 RW 47, 56.}. The foundation of the decision only needs to be comprehensible, not recalculable\footnote{See Hoeren and Niehoff, 2018 \cite{hoeren2018ki} 1 RW 47, 57; and OLG Nürnberg ZD 2013, 26.}. Areas of possible concern when legally assessing the scope of the transparency requirement lie in the application to deep neural networks, where the model is trained to develop its own set of decision-making rules. As these rules are accessible but – for now – not easily interpretable by humans, the relevant fundamentals of the decision-making process cannot be disclosed. 

It becomes clear that this “black box” problem and the way it limits legal compliance in practice had not been considered when the GDPR was brought to life\footnote{See Hoeren and Niehoff, 2018 \cite{hoeren2018ki} 1 RW 47, 58; and DFK Bitkom, 2017 \cite{dfk}.}. Thus, the transparency requirements of the GDPR must be interpreted retrospectively and with regard to the functioning of deep neural networks. In contrast to other state-of-the-art fair representation algorithms, where all underlying rules for a decision remain unrecognisable, the introduction of the presented correction vector in the decision-making process offers an alternate approach to transparency when using deep neural networks. While the evolvement of this set of rules remains part of the ``black box'', the correction vector itself becomes ``human-readable'' -– which may satisfy the GDPR-requirement of disclosing the “logic involved”. This assumption will have to be legally assessed in more detail. 

A second area to be considered in the legal context of our proposal are anti-discrimination rights. In addition to striving for transparency, the introduction of the correction vector aims to meet an even further-reaching fairness concept. Models that learn from past data tend to perpetuate existing structures and thus solidify them. The model approach taken here is to counteract this trend by trying to remove biased information. The general objective is to promote equal opportunities and distributive justice through explicitly penalizing privileged groups on the one hand and by improving unprivileged groups on the other in order to make them more similar and comparable to one another. The way this process is set up may, however, affect anti-discrimination rights. The EU Commission’s recently proposed “first ever legal framework on AI” \cite{euproposal} focuses on the possible impact on fundamental rights when AI technology is applied. The approach of the EU Commission is not to directly deal with the AI techniques as such but rather with their applications and associated risks. Four categories of risk are introduced in the proposal: unacceptable risk, high risk, limited risk, and minimal risk. The principle behind the categories: ``the higher the risk of a specific type of use, the stricter the rules'' \cite{times}. With this proportionate and risk-based approach, the EU aims to create clearer rules for legally compliant and trustworthy use of AI, even if full disclosure of the underlying rules for the decision-making process cannot be ensured \cite{times}. As these recent developments also concern deep neural networks, it will have to be clarified if and how the implementation of our proposed correction vector affects fundamental rights such as anti-discrimination rights.

Notwithstanding these potential legal issues which will have to be determined in more detail, however, by taking into account the need for more transparency while adding fairness aspects into automated decision-making processes through a correction vector, our framework for interpretable fair representation learning should be suitable for contributing to a legally secure and trustworthy use of AI in the EU and inspire further discussion.

\section{Conclusions and Future Work}
\label{sec:conclusions}
In this paper we presented a framework for interpretable fair learning based around the computation of correction vectors. Our experimental results show that existing methodologies may be constrained for explicit computation of correction vectors with negligible losses in performance. Furthermore, our implicit methodology for correction vector computation showed the overall best accuracy/fairness tradeoffs in the considered datasets. While our overall framework is decomposable and allows for inspection of the learned representations, it is still relying on two different black-box components: a future development of our methodology could look into employing neural network models which are themselves explainable. The intuitiveness of the corrections might also benefit from added constraints to their absolute value, so to limit maximum corrections. Another direction for future research could explore extending our framework so that it could handle non-vectorial data such as text. While the models presented in this paper have no specific properties that limit their applicability to tabular data, their architecture has not been optimized for handling other types of data yet. One could explore, as an example, including convolutional layers when analyzing image data or a recurrent mechanism for text. In this field, some authors have focused on learning fair word vectors \cite{zhang2020hurtful} so that generative language models do not display problematic associations such as ``black'' and ``violence''. On the other hand, we are currently not aware of image datasets which lend themselves to group fairness tasks. In this regard, we plan to explore whether the techniques included in this paper might be generalized to interpretable \emph{transformations} of images. Domain adaptation tasks could be the target of this line of research (see, e.g., AlignFlow \cite{alignflow}). Lastly, in the light of recent developments in regulation at the EU level, we reason that our correction vector framework is able to open up the black box of fair DNNs. The legal impact of explicitly computing corrections is, however, not trivial: some individuals belonging to non-protected groups might be explicitly penalized in their features, which might in turn affect anti-discrimination rights. It stands to reason, however, that non-interpretable fair models also compute penalties to complex non-linear combinations of the input features. Our framework, on the other hand, lets human analysts peek inside the ``black box'' of representation learning and see for themselves what the model has learned in a human-readable format. As such, it adds further possibilities for scrutiny of fair models beyond performance on standard metrics, adding to their transparency and trustworthiness.



\printbibliography

\newpage
\section{Supplementary Material}
\label{sec:supplementary}

\subsection{An Introduction to the Real NVP Architecture} \label{sec:nvp_general}
In the following we present a brief introduction to the real NVP architecture \cite{dinh2017realnvp}
with the purpose to make the present paper as self-contained as possible.

Given a $D$-dimensional feature space, the real NVP architecture implements a function $f:\mathds R^D\to\mathds R^D$ and achieves invertibility by employing so-called \textit{affine coupling layers}, which leave part of their input invariant while tranforming the remaining degrees of freedom. To be more specific, for $0<d<D$, $L_s,L_t:\mathds R^d\to\mathds R^{D-d}$,  an affine coupling layer performs the transformation $g:\mathds R^D\to\mathds R^D$ given by
\begin{align}
\begin{aligned}
 x_{1:d}\overset{\text{id}}\mapsto&x_{1:d}, \\
 x_{d+1:D}\mapsto& x_{d+1:D}\odot\exp\left(L_s(x_{1:d})\right)+L_t(x_{1:d}), 
\end{aligned}\label{eq:coupling_layer}
\end{align}
where $\odot$ is the Hadamard product. $s$ and $t$ would usually be implemented by neural network structures of arbitrary kind. This transformation can easily be inverted via the transformation
\begin{align}
\begin{aligned}
 g^{-1}(y)_{1:d}=&y_{1:d}, \\
 g^{-1}(y)_{d+1:D}=&(y_{d+1:D}-L_t(y_{1:d}))\odot\exp\left(-L_s(x_{1:d})\right).
\end{aligned} \label{eq:inverse_coupling_layer}
\end{align}
Since an affine coupling layer only transforms part of its input, a real NVP usually consists of several such layers that transform different components of the input in tandem, allowing to transform all components.

Observing that the transformations (\ref{eq:coupling_layer}) and (\ref{eq:inverse_coupling_layer}) are differentiable, the function $f$ defined by a real NVP is in fact a diffeomorphism. This fact allows for the interpretation of $f$ as a transformation between probability distributions. Given data distributed according to a probability density function (pdf) $P:\mathcal F:[0,\infty)$ on a feature space $\mathcal F\subseteq\mathds R^D$,  the transformed data is distributed according to the pdf $\tilde P:\tilde F\to[0,\infty)$ on the image $\tilde{\mathcal F}\subseteq\mathds R$ of $\mathcal F$ under $f$, given by
\begin{align}
 \tilde P(x)=P\left(f^{-1}(x)\right)\left|\det J_{f^{-1}}(x)\right| \label{eq:pdf_transform}
\end{align}
with $|\det J_{f^{-1}}(x)|$ being the modulus of the Jacobian determinant of $f^{-1}$, which can easily be calculated as the exponential of the sum of all output neurons of all $L_s$-layers within the net.

\subsection{Developing the FairNF Architecture}\label{sec:knn_mixture}
As described extensively in Section~\ref{sec:nf}, pairing two real NVP models might not be enough to obtain fairness. While it is, in principle, possible to transform samples from one distribution to another, the transformation does not remove information about the sensitive data since it is a bijection. 
We recall here our two proposals to solve this issue: (i) Break the bijection $f$ by using a function $f_{pr}$ which sets the first component $z_1$ of the latent feature vector $z \in \mathcal{Z}$ to 0; (ii) Maximize the overlap between $s$ and $z_1$ by employing its value - before deletion - to predict the sensitive attribute via cross-entropy loss. In this section we refer to the base real NVP model as FairNF; the model implementing (i) is referred to as FairNF+$f_{pr}$; the model implementing (ii) is referred to as FairNF+BCE.
\begin{figure}[!ht]
    \centering
    \includegraphics[width=0.49\textwidth]{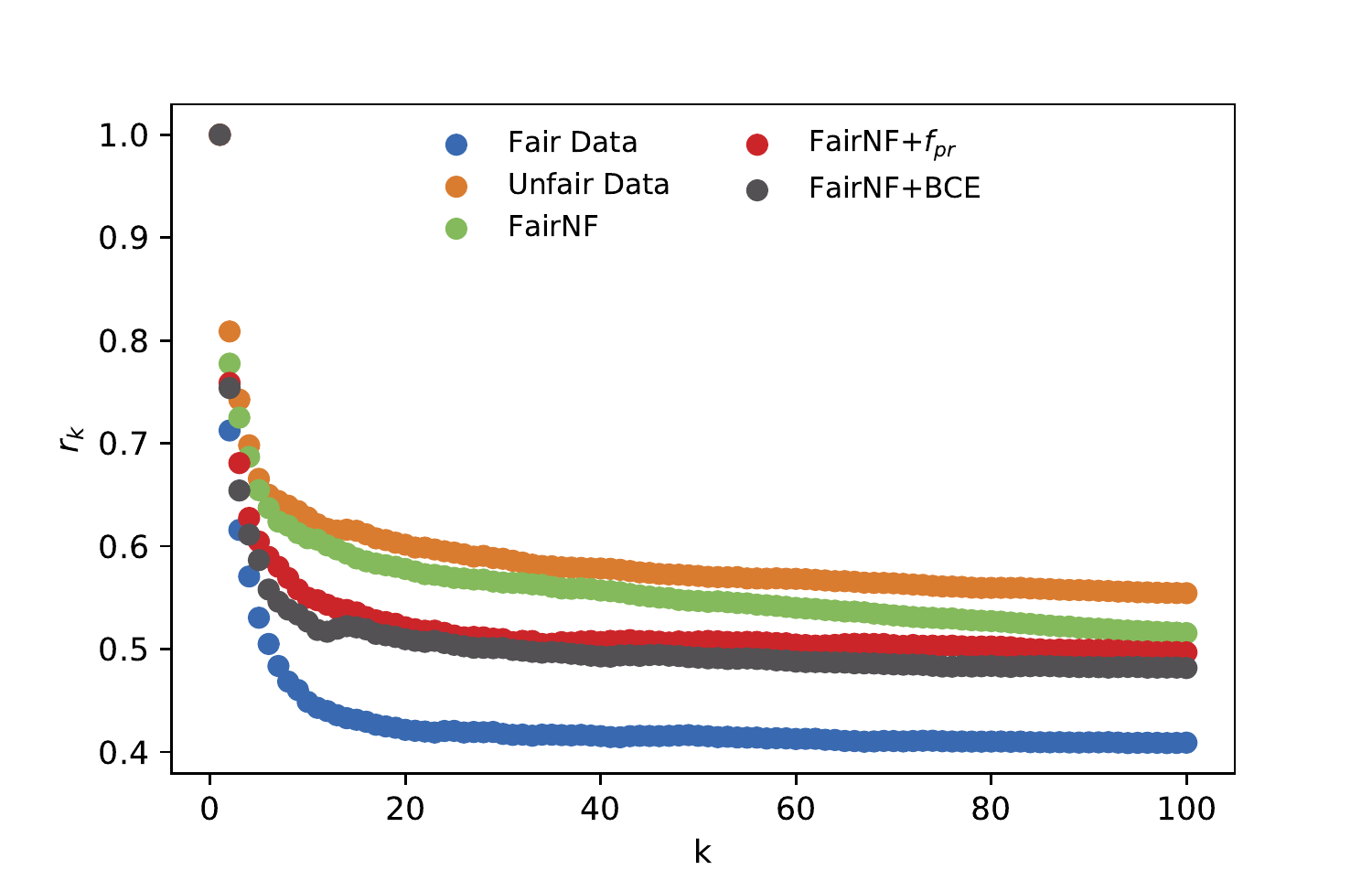}
    \caption{$r_k$ over different $k$ for the three different FairNF models, the fair baseline data and the original unfair data which is transformed by the three models.}
    \label{fig:knn_test}
\end{figure}
To evaluate whether the transformed distributions of the three different FairNF models are mixing the two sensitive groups well enough, we propose the following procedure.
We choose a sensitivity group whose value of the sensitive attribute we take to be $s=0$.
For each instance $x_0\in\mathcal X_0:=\{x\in\mathcal X|s=0\}$ we determine the set of its $k$ nearest neighbors $\text{kNN}(x_0)$ and count the number $n_0(x_0)=\#\{x\in\text{kNN}(x_0)|s=0\}$ of these nearest neighbors belonging to the same sensitivity group as $x_0$.
Furthermore we calculate the average ratio 

\begin{equation}
    r_k=\frac1{\#\mathcal X_0}\sum_{x_0\in\mathcal X_0}\frac{n_0(x_0)}k.
\end{equation}
In the case of a well mixed dataset this ratio should converge to the overall dataset ratio of $\hat r=\frac{\#\mathcal X_0}{\#\mathcal X}$ when increasing $k$.
From this, we define a mixture metric by averaging $r_k$ over different values for $k$ up to a predefined threshold $T$:
\begin{equation}
    M=1-\frac1{T\hat r}\sum_{k=1}^Tr_k.
\end{equation}
For evaluating the proposed metric and quantify which of the three FairNF algorithms has the better mixing of the target data, we generate two sets of synthetic data by randomly sampling two features from different normal distributions.
In the first case both sensitive attributes of the generated data have the same mean and the standard deviation.
In the second case we choose different means for the two distributions, representing two separate groups with different values of a sensitive attribute ($\mu_0 = [0,0]$ and $\mu_1 = [1,1]$).
We consider the first synthetic data as well mixed and we will use this in the following as a fair baseline, while the second dataset is considered to be unfair and will be further transformed by the three different kind of FairNF models.
We plot in Figure~\ref{fig:knn_test} $r_k$ over different $k$-NN values.
It is clear that the value for FairNF+BCE is closer to the fair baseline, while the other two FairNF models are not closer to the unfair dataset.
For higher $k$ the FairNF approach is lowering $r_k$ under the values of the FairNF+$f_{pr}$ model.
This behavior shows that naively removing one component of the latent space does not result in fairer results.
Nevertheless the approach of the FairNF+BCE model seems to be able to mix the sensitive attributes in the target space well.

\begin{figure*}[!ht]
\centering
\subfigure[Fair Synthetic Data]{\label{fig:toy_1}\includegraphics[width=0.49\textwidth]{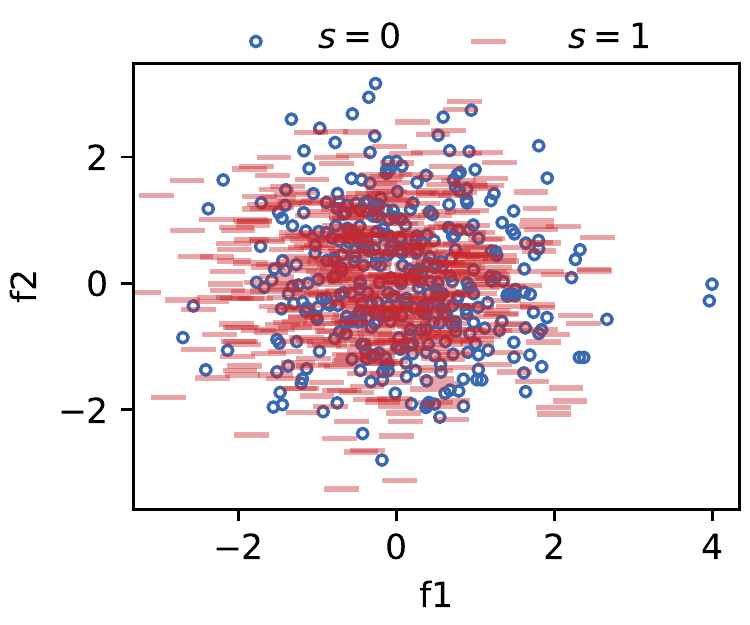}}
\subfigure[Unfair Synthetic Data]{\label{fig:toy_2}\includegraphics[width=0.49\textwidth]{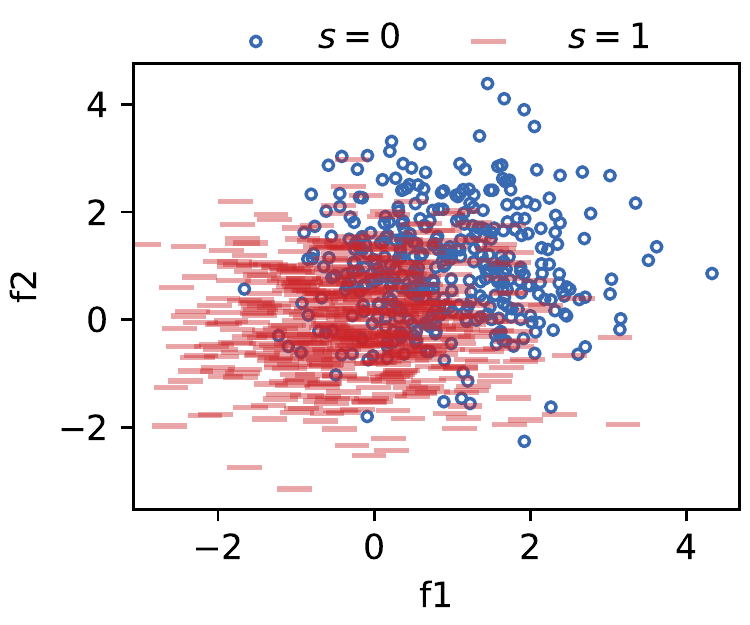}}
\subfigure[FairNF]{\label{fig:toy_3}\includegraphics[width=0.32\textwidth]{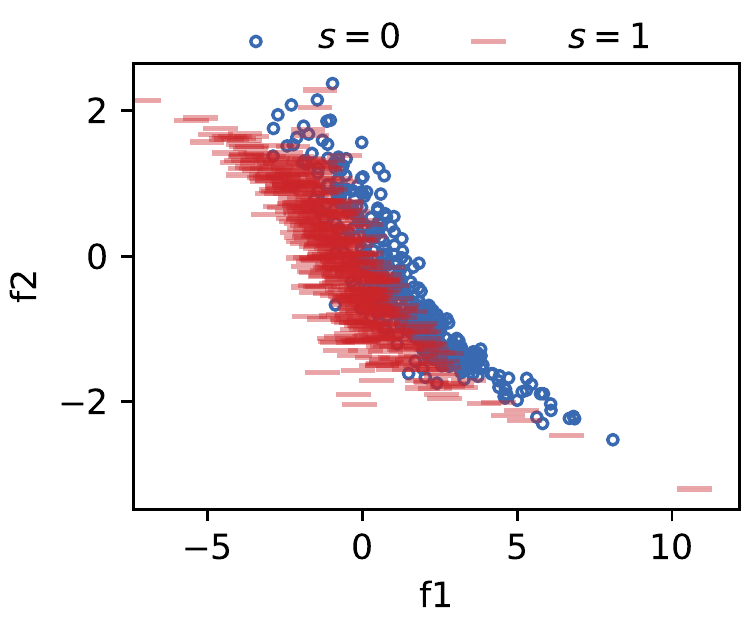}}
\subfigure[FairNF+$f_{\text{pr}}$]{\label{fig:toy_4}\includegraphics[width=0.32\textwidth]{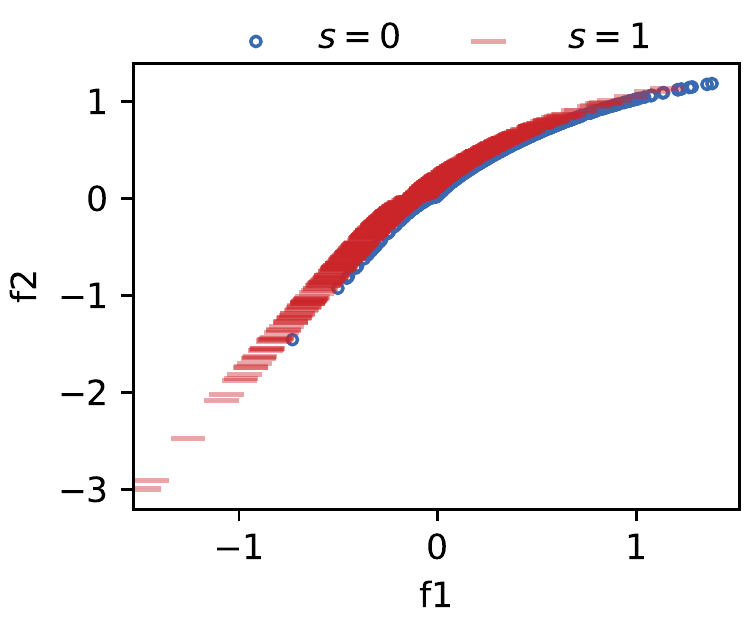}}
\subfigure[FairNF+BCE]{\label{fig:toy_5}\includegraphics[width=0.32\textwidth]{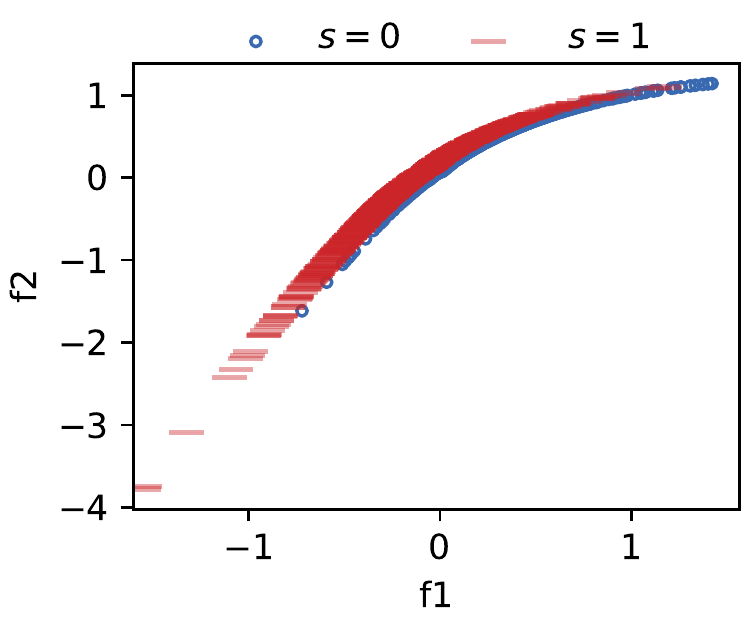}}
\caption{In~\ref{fig:toy_1} the generated fair data is shown where the different groups are displayed with blue and red. In~\ref{fig:toy_2} the generated unfair data is shown. \ref{fig:toy_3}, \ref{fig:toy_4} and \ref{fig:toy_5} show the generated unfair data using the FairNF, the FairNF+$f_{\text{pr}}$ and FairNF+BCE algorithm.}\label{fig:toy_all}
\end{figure*}
A representation of how this simple two-Gaussians dataset is transformed after application of the three different FairNF models may be found in Figure~\ref{fig:toy_all}. The axes display feature values. In Figure \ref{fig:toy_1} the fair dataset is shown while Figure \ref{fig:toy_2} shows the unfair data which will be transformed.
In Figure \ref{fig:toy_3}, Figure \ref{fig:toy_4} and  Figure \ref{fig:toy_5} the transformed data is shown.
The first one shows the data transformed by the FairNF algorithm.
The second figure was transformed using the FairNF+$f_{pr}$ while the third figures shows the result of the FairNF+BCE model.
For both the FairNF+$f_{pr}$ and the FairNF+BCE, we observe a strong mixing of the two sensitivity groups, wheras the base FairNF still keeps the groups apart in the latent representation. After gathering this evidence, we decided to employ the FairNF+BCE model in all our experiments, which are reported in Section~\ref{sec:experiments} in the main body of the paper.

\subsection{Datasets} \label{sec:app_datasets}
In the following we give an extended description of the four datasets we employed in our experimentation. We also clarify how they might be employed in classification and ranking settings. 
The COMPAS dataset was released as part of an investigative journalism effort in tackling automated discrimination \cite{machine_bias}. We employed the original risk score in our ranking experiments; and indicator variable representing whether an individual committed a crime in the following two years was employed in our classification experiments.
In terms of correction vectors analysis (see Section \ref{sec:cor_ana}) we evaluated the feature ``prior crimes'' and how much it was changed by the interpretable models to make the representations fair.
Moreover, the Adult dataset is used, where the ground truth represents whether an individual's annual salary is over 50K\$ per year or not \cite{adult}. 
It is commonly used in fair classification, since it is biased against gender \cite{Louizos2016TheVF,zemel2013learning,cerrato2020constraining}.
For the correction vectors analysis, the feature ``capital gain'' was analyzed.
The third dataset used in our experiments is the Bank Marketing Data Set \cite{banks}, where the classification goal is whether a client will subscribe a term deposit.
The dataset is biased against people under 25 or over 65 years.
The ``employment variation rate'' was studied for the correction vectors analysis.
The last dataset we used is the Law Students dataset, which contains information relating to 21,792 US-based, first-year law students~\cite{law_student}. As done previously~\cite{zehlike2018reducing}, we subsampled 10\% of the total samples while maintaining the distribution of gender and ethnicity. We used ethnicity as the sensitive attribute while sorting students based on their predicted academic performance.
Here, we analysed the change in LSAT score during our correction vectors analysis.

\subsection{Metrics}\label{sec:metrics-appendix}

\subsubsection{Ranking Metrics}\label{sec:rank-metric}

\paragraph*{The NDCG Metric.}\label{sec:app_ndcg}\ The normalized discounted cumulative gain of top-$k$ documents retrieved (or NDCG@$k$) is a commonly used measure for performance in the field of learning to rank.
Based on the cumulative gain of top-$k$ documents (DCG@$k$), the NDCG@$k$ can be computed by dividing the DCG@$k$ by the ideal (maximum) discounted cumulative gain of top-$k$ documents retrieved (IDCG@$k$):
\begin{equation}
\text{NDCG@}k = \frac{\text{DCG@}k}{\text{IDCG@}k} = \frac{\sum_{i=1}^{k} \frac{2^{r(d_i)} - 1}{\log_2(i + 1)}}{\text{IDCG@}k}
\,,\notag
\end{equation}
where $d_1, d_2, ..., d_n$ is the list of documents sorted by the model with respect to a single query and $r(d_i)$ is the relevance label of document $d_i$.

\paragraph*{rND.}\label{sec:app_rnd}\ To the end of measuring fairness in our models, we employ the rND metric~\cite{ke2017measuring}. 
This metric is used to measure group fairness and is defined as follows: 

\begin{equation}
\text{rND} = \frac{1}{Z} \sum_{i \in \{10, 20, ...\}}^N \frac{1}{\log_{2}(i)} \left| \frac{ \mid S^{+}_{1...i} \mid}{i} - \frac{\mid S^+ \mid}{N} \right |.
\end{equation}

The goal of this metric is to measure the difference between the ratio of the protected group in the top-$i$ documents and in the overall population.
The maximum value of this metric is given by $Z$,  which is also used as normalization factor.
This value is computed by evaluating the metric with a dummy list, where the protected group is placed at the end of the list. This biased ordering represents the situation of ``maximal discrimination''. 

This metric also penalizes if protected individuals at the top of the list are over-represented compared to their overall representation in the population.

\paragraph*{Group-dependent Pairwise Accuracy.}\label{sec:app_gpa}\ Let $G_1, ..., G_K$ be a set of $K$ protected groups such that every instance inside the dataset $\mathfrak D$ belongs to one of these groups. The \emph{group-dependent pairwise accuracy} \cite{fair_pair_metric} $A_{G_i > G_j}$ is then defined as the accuracy of a ranker on instances which are labeled more relevant belonging to group $G_i$ and instances labeled less relevant belonging to group $G_j$. Since a fair ranker should not discriminate against protected groups, the difference $|A_{G_i > G_j} - A_{G_j > G_i}|$ should be close to zero. In the following, we call the Group-dependent Pairwise Accuracy {\em GPA}. We also note that this metric may be employed in classification experiments by considering a classifier's accuracy when computing $A_{G_i}$ and $A_{G_j}$. 

\subsubsection{Classification Metrics}\label{sec:cls-metric}

\paragraph*{Area Under Discrimination Curve.}\label{sec:app_AUDC}\ We take the discrimination as a measure of the bias with respect to the sensitive feature $s$ in the classification \cite{zemel2013learning}, which is given by:
\begin{equation}
\text{yDiscrim} = \left | \frac{\sum^n_{n:s_n=1} \hat{y}_n}{\sum^n_{n:s_n=1} 1} - \frac{\sum^n_{n:s_n=0} \hat{y}_n}{\sum^n_{n:s_n=0} 1} \right |, 
\end{equation}
where $n:s_n=1$ denotes that the $n$-th example has a value of $s$ equal to 1. This measure can be seen as a statistical parity which measures the difference of the proportion of the two different groups for a positive classification.
Similar to the AUC we can evaluate this measure for different classification thresholds and calculate the area under this curve.
Using different thresholds, dependencies on the yDiscrim measure can be taken into account.
We call this metric in the following AUDC (Area Under the Discrimination Curve). 
One issue with this metric is that it may hide high discrimination values on certain thresholds as they will be ``averaged away''. We show plots analyzing the accuracy/discrimination of our models tradeoff at various thresholds in Figure~\ref{fig:compare_auc}. These were obtained by computing the discrimination and accuracy of our models at 20 different thresholds on the interval $[0.05, 1)$. Overall we find that our models find sensible fairness/accuracy tradeoffs for all the thresholds we considered.

\paragraph*{Absolute Distance to Random Guess.}\label{sec:app_ADRG}\ To evaluate the invariance of the representation with respect to the sensitive attribute, we report classifier accuracy as the absolute distance from a random guess (the majority class ratio in the dataset), which we call ADRG in the following.

\subsection{Statistical Comparison of Interpretable vs. Non-Interpretable Models}\label{sec:app_compare_model}
\begin{figure*}[!ht]
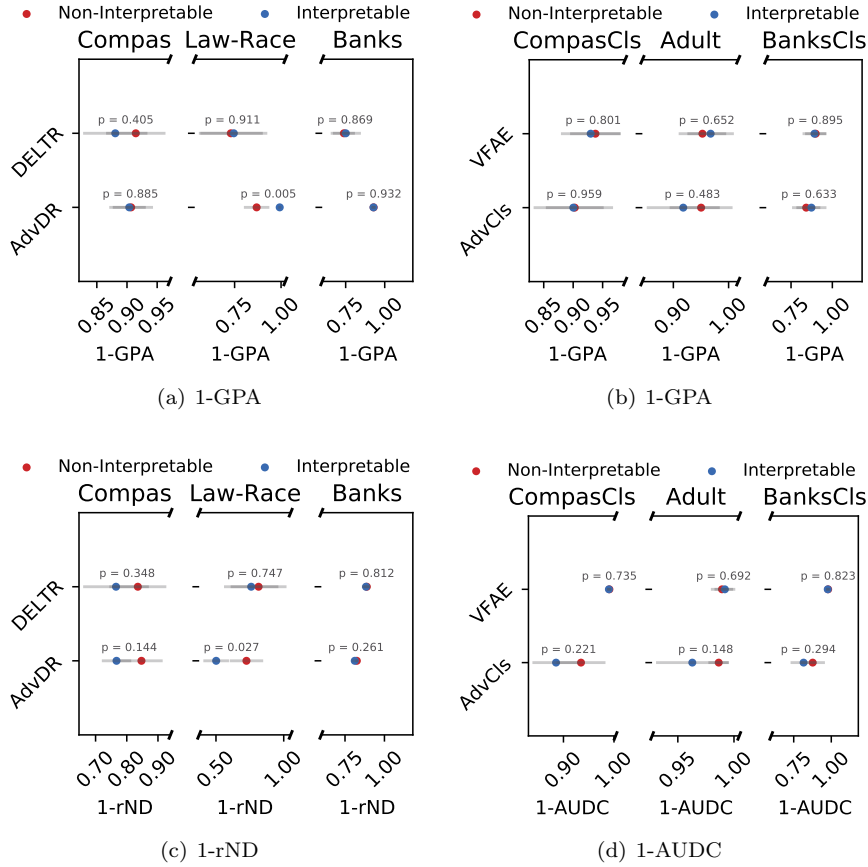

\centering
\subfigure[1-GPA]{\label{fig:app_compare_gpa_ranking}\includegraphics[width=0.49\textwidth]{img/compare_int_1-GPA_ranking.pdf}}
\subfigure[1-GPA]{\label{fig:app_compare_gpa}\includegraphics[width=0.49\textwidth]{img/compare_int_1-GPA.pdf}}
\subfigure[1-rND]{\label{fig:app_compare_rnd}\includegraphics[width=0.49\textwidth]{img/compare_int_1-rND.pdf}}
\subfigure[1-AUDC]{\label{fig:app_compare_audc}\includegraphics[width=0.49\textwidth]{img/compare_int_1-AUDC.pdf}}
\caption{Results for all datasets comparing the fairness of each model with and without constraining it to be interpretable. The blue points represents all interpretable models, while the red points show the non interpretable models. The grey lines around each model represent the error computed over 3 folds.}
\label{fig:interpretable-comparison}
\end{figure*}

To the end of understanding the \emph{fairness} impact of the architectural constraints we imposed to allow for explicit computation of fairness vectors (Section~\ref{sec:interpretable-net}) we perform further experimentation. Each of the architectures that were constrained for direct computation of correction vectors were compared to their non-interpretable versions trained as described in the respective papers introducing them. We trained each interpretable/non-interpretable pair of architectures 100 times while searching for the best hyperparameters. Each dataset is split into 3 internal folds which are employed for hyperparameter selection purposes, while 15 external folds are kept separate until test time. We then analyze the average performance on the relevant fairness metrics (1-GPA, 1-rND for ranking; 1-AUDC for classification) on the external folds. We correct the variance estimation as described by Nadeau and Bengio \cite{nadeau2003inference} and perform a significance t-test. Here the number of runs averaged is the number of test (external) folds $n=15$ following the recommendations by Nadeau and Bengio, who found little benefit in higher values in terms of statistical power. 

We report the pairwise comparisons between interpretable and non-interpretable models in Figure~\ref{fig:interpretable-comparison}. Each comparison also reports the p-value obtained by the application of the t-test. We found that the data does not support the presence of statistically significant differences between the fairness performance of our interpretable models and the non-interpretable ones. One model/dataset combination does take exception to this, however. This exception is represented by the Law-Race dataset when learning AdvDR ranking models. In Figure~\ref{fig:app_compare_gpa} we observe a statistically significant advantage in GPA when employing the interpretable model, whereas this trend is reversed when considering rND (Figure~\ref{fig:app_compare_rnd}). We hypothesize here that our hyperparameter search, which seeks to find models with the best values of $(1-GPA) + (1-rND)$, has found an impressively fair interpretable model for 1-GPA while sacrificing rND performance. The interpretable AdvDR does not seem to suffer much in terms of rND on other datasets, which suggests that it has no issue optimizing rND in general. We conclude that our framework for explicit computation of correction vectors warrants strong consideration in situations where transparency is needed to comply with existing laws and regulations.

\subsection{Correction Vectors Analysis}

\subsubsection{Synthetic data}\label{sec:cor_ana_toy}
To provide a more visual example, we provide here pre- and post-correction visualizations for a simple two-Gaussians dataset, which can be found in Figure~\ref{fig:direction_plots}. This dataset consists of samples from two separate Gaussian distributions which only differ in their mean. More specifically, the mean vectors are $\mu_0 = [0, 0]$ and $\mu_1 = [1, 1]$, forming two point clouds which partially overlap as we opted for an unit-valued variance for both. We employ this dataset in training an interpretable AdvCls \cite{cerrato2020constraining,xie2017controllable}, which performs explicit corrections (see Section~\ref{sec:interpretable-net}), and a FairNF model (see Section~\ref{sec:nf}). What we observe in the plots is the AdvCls model performing rather intuitive translations, which push the two Gaussian distributions closer in feature space after correction. The FairNF model instead opts to push both distributions into a line-like manifold. We reason that this is a side effect of our strategy to break the bijection feature of Normalizing Flows via discarding one of the features in the latent space $\mathcal Z$. We do note that the examples are centered around $[0, 0]$ after correction, meaning that the second Gaussian has been pushed onto the first. It is noteworthy, however, that the original distribution shapes were not kept equal, even if the variance appears to be similar.

\begin{figure*}[!ht]
\centering
\subfigure[Direction Plot IntAdvCls]{\label{fig:dir_1}\includegraphics[width=0.49\textwidth]{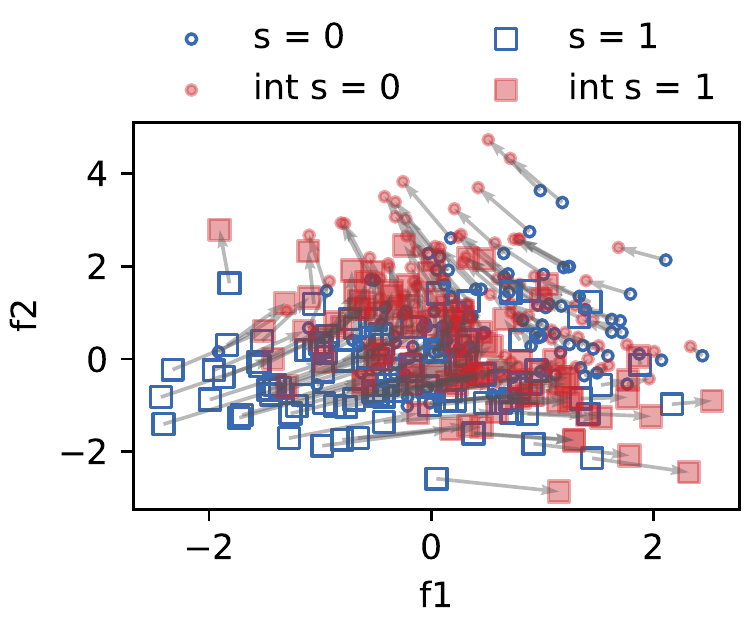}}
\subfigure[Direction Plot FairNF]{\label{fig:dir_2}\includegraphics[width=0.49\textwidth]{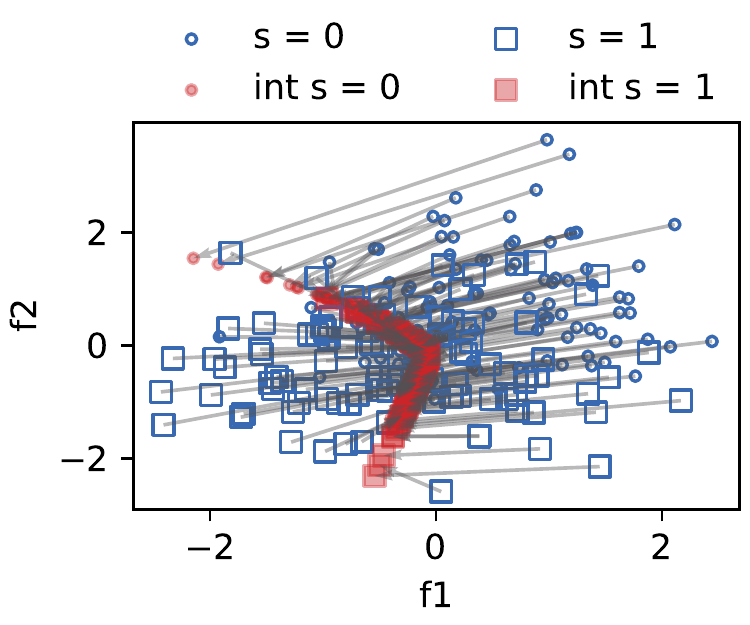}}
\caption{Direction plots for the computed corrections in a AdvCls and FairNF model trained on a two-Gaussians dataset. The two distributions have unit variance and differ only in their mean. Each distribution is akin to a different value of a binary-valued sensitive attribute, with circles representing $s=0$ and squares representing $s=1$. The original samples are plotted in \textcolor{blue}{blue}; the corrected samples are plotted in \textcolor{red}{red}.} \label{fig:direction_plots}
\end{figure*}

\subsubsection{Real world data}\label{sec:cor_ana_exp}

\begin{figure*}[!ht]
\centering
\subfigure[Correction Vector Ranking]{\label{fig:cor_ranking}\includegraphics[width=0.98\textwidth]{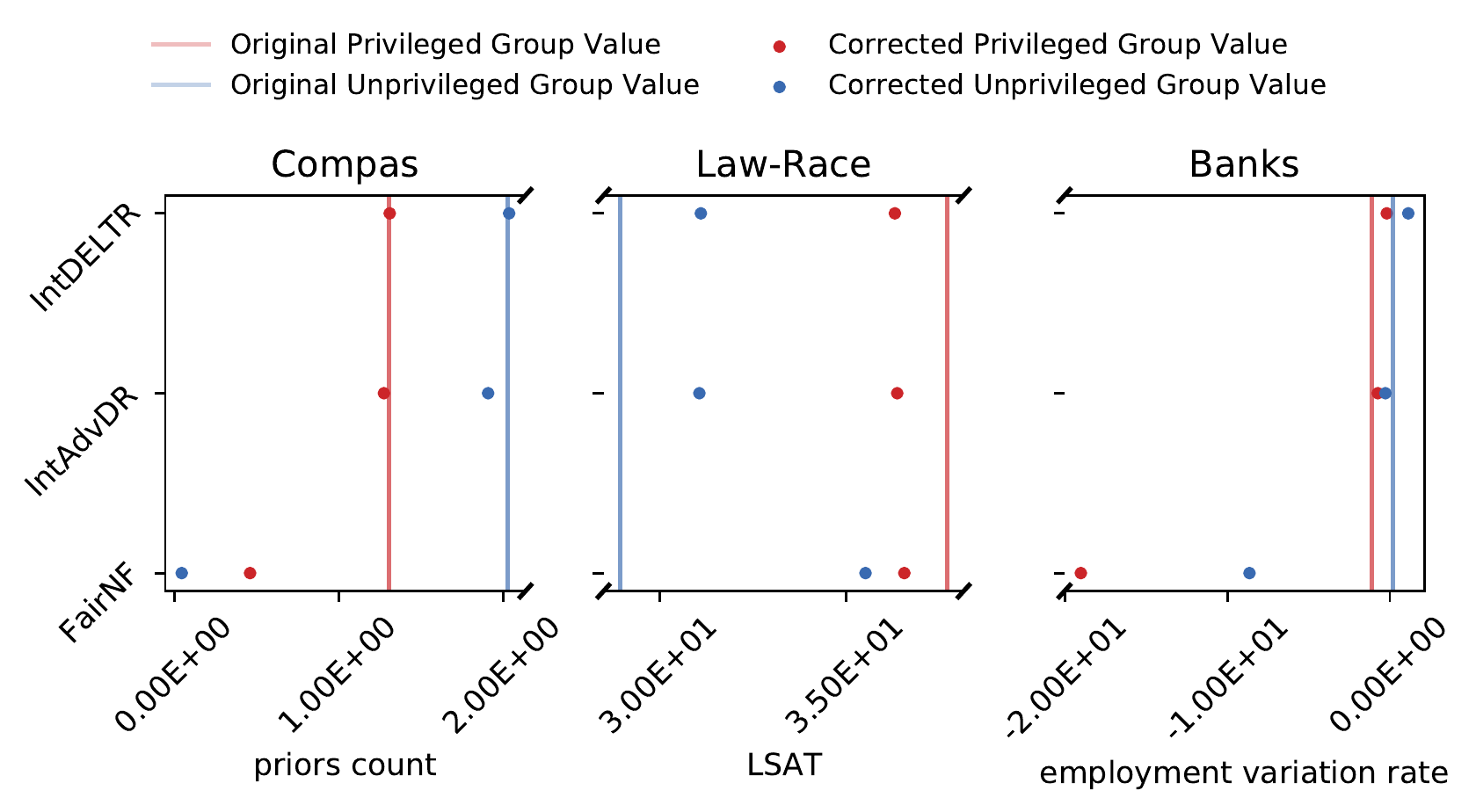}}
\subfigure[Correction Vector Classification]{\label{fig:cor_cls}\includegraphics[width=0.98\textwidth]{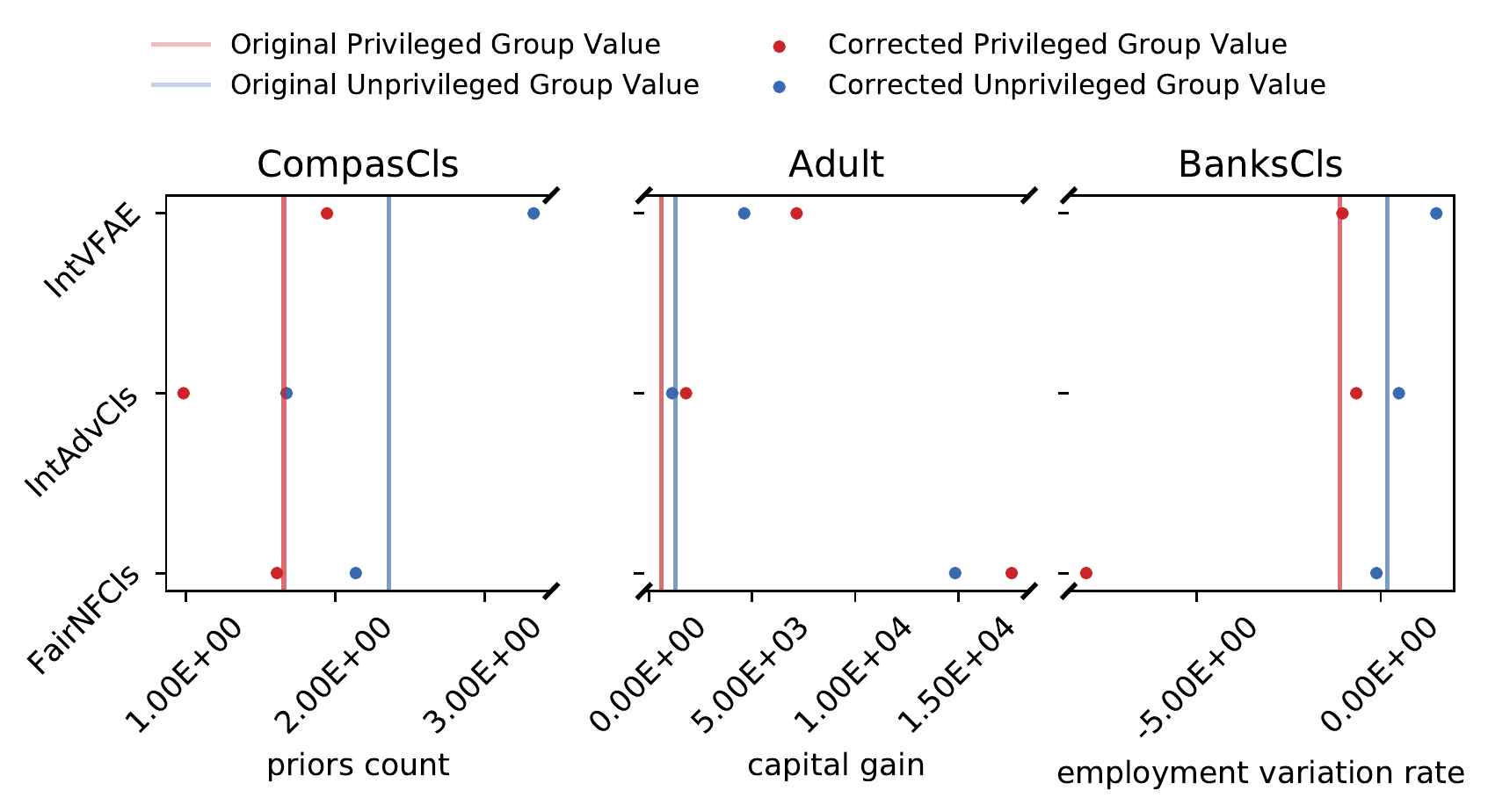}}
\caption{Average corrections for the best performing models over all datasets. We focus on one feature for which we have an expectation that the average difference should be reduced after the fair correction.}\label{fig:interpretable-vects}
\end{figure*}

After having built a visual intuition about how explicit and implicit corrections might look and differ, we now move on to analyzing the corrections on the real-world datasets described in Section~\ref{sec:datasets}. Our analysis was performed as follows. For each dataset, we pick a feature for which we expect correction vectors to make individuals belonging to different groups more similar. We then compute the average correction for each group and compute the average difference in the chosen feature in absolute and relative terms. A figure with the average corrections for the best performing models over all datasets is given in Figure~\ref{fig:interpretable-vects}.
A table with all average corrections for each model and dataset can be found in Table~\ref{tab:interpretable-vects}. 
What stands out from these results is that different models often disagree about both the sign and the intensity of the correction that should be applied. For instance, on the COMPAS dataset, we would expect a correction to reduce the difference in the average amount of previous crimes committed between white and black people. While our implicit computation methodology does just that, the VFAE constrained for explicit computation of correction vector opts to widen that gap. Nonetheless, it obtains an overall fair result (see Figures~\ref{fig:compare_audc} and \ref{fig:compare_gpa}). Similar extreme corrections, which are definitely not intuitive, were computed by the interpretable AdvCls and FairNF models on the Adult dataset. One noteworthy trend is that simpler architectures computed more intuitive corrections. To be more specific, both the AdvDR and DELTR models employ only a single neural layer after computing the corrected features. In both models this layer has a single neuron which outputs the ranking scores ($p(y_1 > y_2)$ for the pairwise AdvDR and $p_{top_1}(x_i)$ for the listwise strategy DELTR). This severely constrains the complexity of the transformation learned by the models after feature vector computation and is highly in contrast with the other models, which have no such restrictions. DELTR and AdvDR are able to clearly close the gap in average feature values in all considered datasets.
The other models were also able to output reasonable corrections. For instance, our FairNF model opted to almost nullify the gap in prior crimes committed on the COMPAS dataset. However, these models appear to be less consistent in this regard -- FairNF instead opted to amplify the average difference in employment variation rate on the Adult dataset. Interestingly, this model is the strongest tradeoff performer on that same dataset (Figure~\ref{fig:compare_gpa} and~\ref{fig:compare_audc}). As previously discussed, we hypothesize that the absence of restrictions about how many layers may be included after the computation of the correction vectors may limit the intuitiveness of some of the corrections. This issue could be addressed in the future by further constraining the maximum computable value for correction vectors.
These matters notwithstanding, the opportunity to peek inside the ``black box'' of fair representation learning might also guide how a practitioner would select models: a model which computes corrections that are deemed counter-intuitive or excessive may be discarded during model selection. Therefore, our framework adds a new layer of opportunities in developing models that can be trustworthy and transparent on top of fair. 

We report in Figure~\ref{tab:interpretable-vects} a table displaying the average correction vectors for all datasets and models considered. These models were obtained via hyperparameter selection as described in Section~\ref{sec:experiments}. We take a single feature from each dataset for which we expect our models to close the gap between historically privileged and unpriviledged groups. We observe that simpler models, which include only a single layer after correction vector computation, display more intuitive corrections, whereas even very fair models might learn corrections that are not as easy to understand. We refer to Section~\ref{sec:cor_ana} for an in-depth discussion of these matters. 

\begin{figure*}[!ht]
\centering
\begin{adjustbox}{angle=90, width=\textwidth}
\begin{tabular}{lllllll}
\hline
  Dataset &     Model &                   Feature &  Original &       Privileged Group &    Unprivileged Group &       Avg. Difference \\
\hline
   Compas &    FairNF &              priors count &  original &                  1.854 &                 2.559 &                -0.704 \\
   Compas &    FairNF &              priors count & corrected &         0.92 (-50.4\%) &       0.902 (-64.8\%) &      0.018 (-102.6\%) \\
 Law-Race &    FairNF &                      LSAT &  original &                 37.709 &                28.928 &                 8.781 \\
 Law-Race &    FairNF &                      LSAT & corrected &         39.882 (5.8\%) &       50.553 (74.8\%) &    -10.671 (-221.5\%) \\
    Banks &    FairNF & employment variation rate &  original &                 -1.108 &                 0.171 &                -1.279 \\
    Banks &    FairNF & employment variation rate & corrected &      -11.815 (966.2\%) &    -2.592 (-1614.6\%) &      -9.223 (621.0\%) \\
   Compas &  IntAdvDR &              priors count &  original &                  1.854 &                 2.559 &                -0.704 \\
   Compas &  IntAdvDR &              priors count & corrected &         1.799 (-3.0\%) &        2.503 (-2.2\%) &        -0.705 (0.1\%) \\
 Law-Race &  IntAdvDR &                      LSAT &  original &                 37.709 &                28.928 &                 8.781 \\
 Law-Race &  IntAdvDR &                      LSAT & corrected &       33.102 (-12.2\%) &        30.913 (6.9\%) &       2.189 (-75.1\%) \\
    Banks &  IntAdvDR & employment variation rate &  original &                 -1.108 &                 0.171 &                -1.279 \\
    Banks &  IntAdvDR & employment variation rate & corrected &        -0.99 (-10.7\%) &     -0.261 (-252.8\%) &      -0.728 (-43.1\%) \\
   Compas &  IntDELTR &              priors count &  original &                  1.854 &                 2.559 &                -0.704 \\
   Compas &  IntDELTR &              priors count & corrected &         1.685 (-9.2\%) &        2.373 (-7.3\%) &       -0.688 (-2.4\%) \\
 Law-Race &  IntDELTR &                      LSAT &  original &                 37.709 &                28.928 &                 8.781 \\
 Law-Race &  IntDELTR &                      LSAT & corrected &        37.317 (-1.0\%) &       32.105 (11.0\%) &       5.213 (-40.6\%) \\
    Banks &  IntDELTR & employment variation rate &  original &                 -1.108 &                 0.171 &                -1.279 \\
    Banks &  IntDELTR & employment variation rate & corrected &       -0.595 (-46.3\%) &        0.58 (238.8\%) &       -1.175 (-8.1\%) \\
CompasCls & FairNFCls &              priors count &  original &                  1.854 &                 2.559 &                -0.704 \\
CompasCls & FairNFCls &              priors count & corrected &        0.861 (-53.6\%) &       0.871 (-66.0\%) &       -0.01 (-98.6\%) \\
    Adult & FairNFCls &              capital gain &  original &                610.118 &              1292.517 &              -682.399 \\
    Adult & FairNFCls &              capital gain & corrected &     6370.103 (944.1\%) &    37664.4 (2814.0\%) & -31294.297 (4485.9\%) \\
 BanksCls & FairNFCls & employment variation rate &  original &                 -1.108 &                 0.171 &                -1.279 \\
 BanksCls & FairNFCls & employment variation rate & corrected &       -5.504 (396.8\%) &    -2.373 (-1486.7\%) &      -3.132 (144.8\%) \\
CompasCls & IntAdvCls &              priors count &  original &                  1.854 &                 2.559 &                -0.704 \\
CompasCls & IntAdvCls &              priors count & corrected &         1.775 (-4.3\%) &        2.455 (-4.0\%) &        -0.68 (-3.5\%) \\
    Adult & IntAdvCls &              capital gain &  original &                610.118 &              1292.517 &              -682.399 \\
    Adult & IntAdvCls &              capital gain & corrected &    -1873.36 (-407.0\%) &   -403.821 (-131.2\%) &    -1469.54 (115.3\%) \\
 BanksCls & IntAdvCls & employment variation rate &  original &                 -1.108 &                 0.171 &                -1.279 \\
 BanksCls & IntAdvCls & employment variation rate & corrected &        -1.456 (31.4\%) &     -0.567 (-431.7\%) &      -0.889 (-30.5\%) \\
CompasCls &   IntVFAE &              priors count &  original &                  1.854 &                 2.559 &                -0.704 \\
CompasCls &   IntVFAE &              priors count & corrected &         2.181 (17.6\%) &        3.605 (40.9\%) &      -1.424 (102.2\%) \\
    Adult &   IntVFAE &              capital gain &  original &                610.118 &              1292.517 &              -682.399 \\
    Adult &   IntVFAE &              capital gain & corrected & -11422.525 (-1972.2\%) & -10696.029 (-927.5\%) &      -726.496 (6.5\%) \\
 BanksCls &   IntVFAE & employment variation rate &  original &                 -1.108 &                 0.171 &                -1.279 \\
 BanksCls &   IntVFAE & employment variation rate & corrected &       -2.606 (135.2\%) &        0.13 (-24.0\%) &      -2.736 (113.9\%) \\
\hline
\end{tabular}
\end{adjustbox}
\caption{Average corrections for the best performing models over all datasets. We focus on one feature for which we have an expectation that the average difference should be reduced after the fair correction.}\label{tab:interpretable-vects}
\end{figure*}

\subsection{Fair Representations}\label{sec:fair_rep}

\begin{figure*}[!ht]
\centering
\subfigure[Compas Ranking]{\label{fig:ranking_compas_auc}\includegraphics[width=0.32\textwidth]{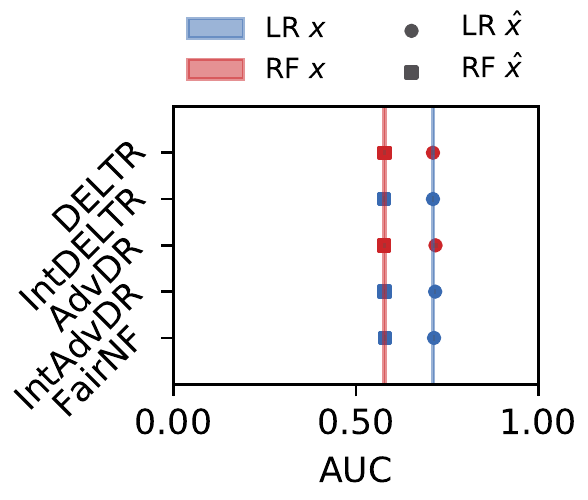}}
\subfigure[Law-Race Ranking]{\label{fig:ranking_law_auc}\includegraphics[width=0.32\textwidth]{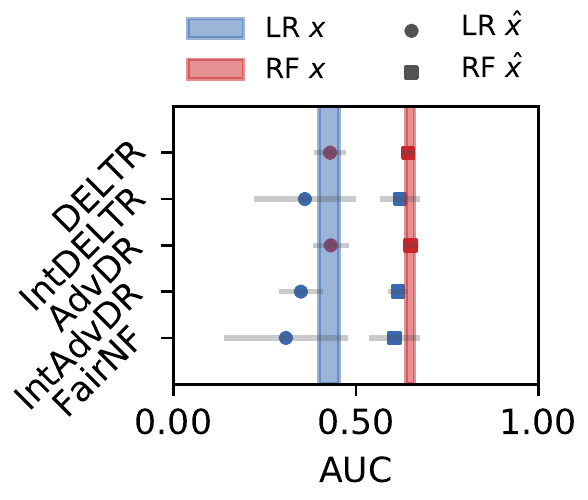}}
\subfigure[Banks Ranking]{\label{fig:ranking_banks_auc}\includegraphics[width=0.32\textwidth]{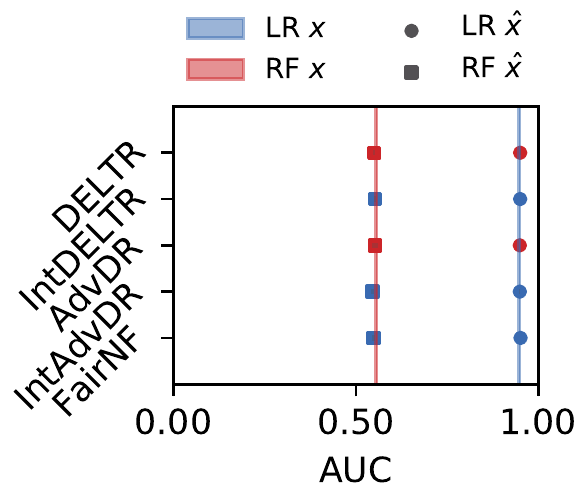}}
\subfigure[Compas Ranking]{\label{fig:ranking_compas_adrg}\includegraphics[width=0.32\textwidth]{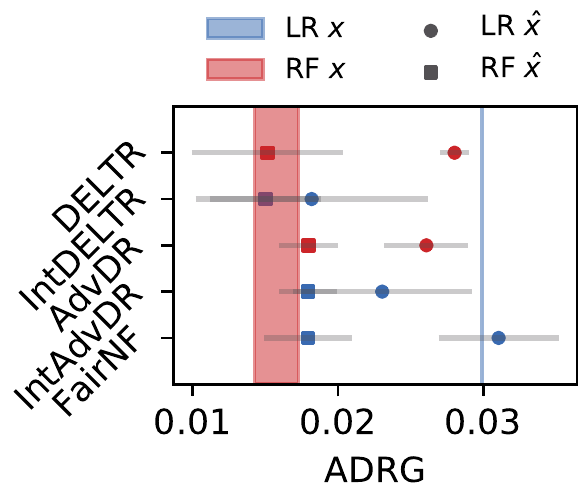}}
\subfigure[Law-Race Ranking]{\label{fig:ranking_law_adrg}\includegraphics[width=0.32\textwidth]{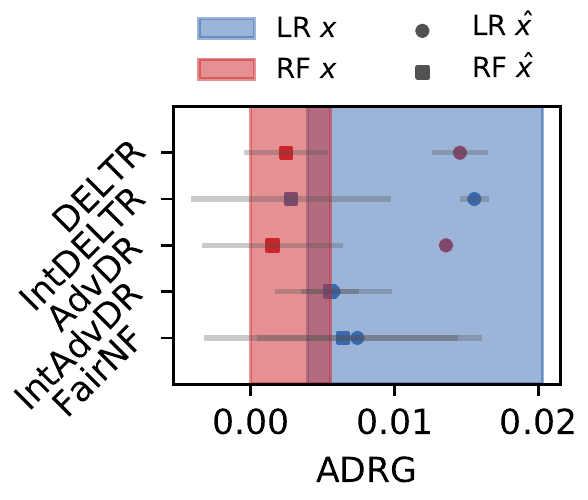}}
\subfigure[Banks Ranking]{\label{fig:ranking_banks_adrg}\includegraphics[width=0.32\textwidth]{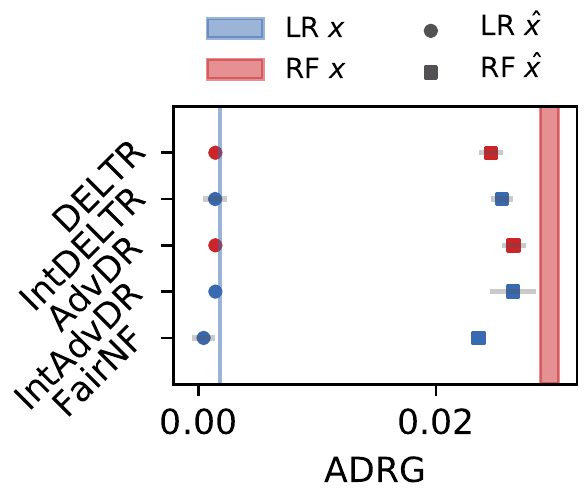}}
\caption{Performance of predicting the sensitive attribute of the external rankers trained on the representations of the different models. The lines show the baseline on the untransformed dataset $x$, while the points show the different results on the model representations $\hat{x}$. Note that despite the ADRG metric having an asymmetric distribution, we employed a symmetric error computation.}\label{fig:ranking_reps}
\end{figure*}

\begin{figure*}[!ht]
\centering
\subfigure[Compas Classification]{\label{fig:cls_compas_auc}\includegraphics[width=0.32\textwidth]{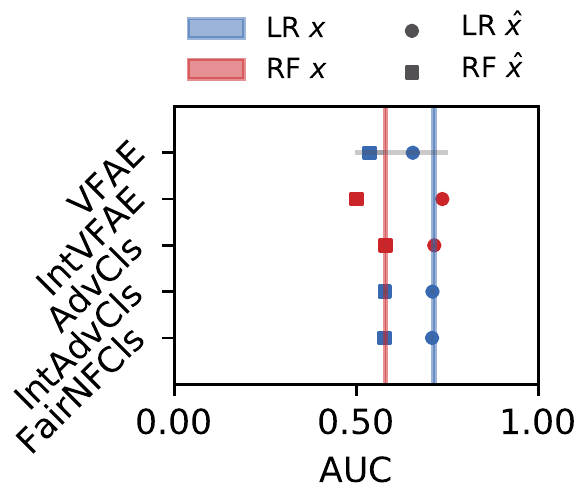}}
\subfigure[Adult Classification]{\label{fig:cls_adult_auc}\includegraphics[width=0.32\textwidth]{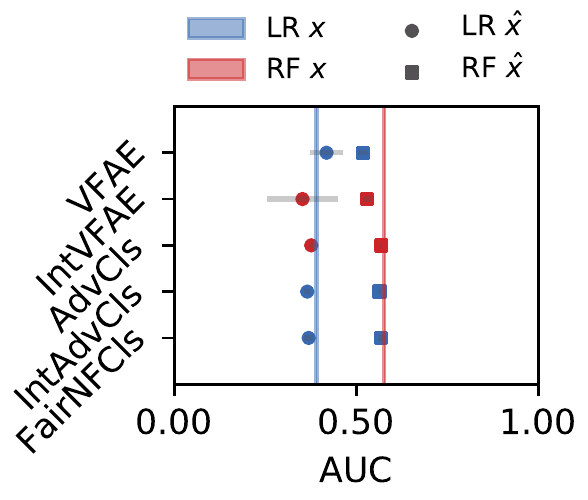}}
\subfigure[Banks Classification]{\label{fig:cls_banks_auc}\includegraphics[width=0.32\textwidth]{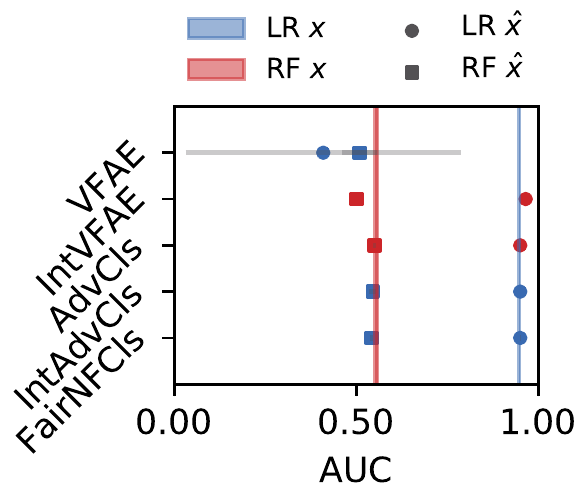}}
\subfigure[Compas Classification]{\label{fig:cls_compas_adrg}\includegraphics[width=0.32\textwidth]{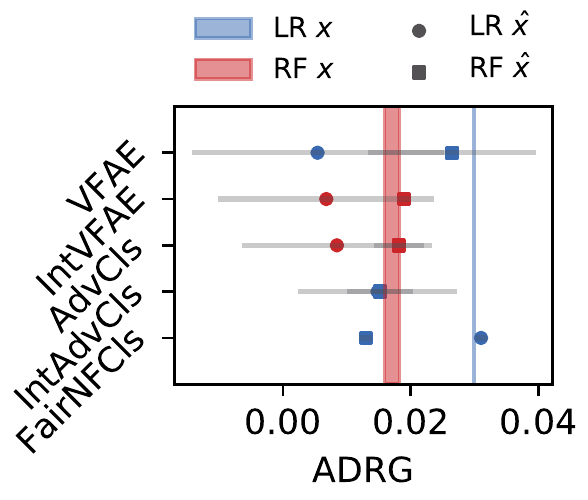}}
\subfigure[Adult Classification]{\label{fig:cls_adult_adrg}\includegraphics[width=0.32\textwidth]{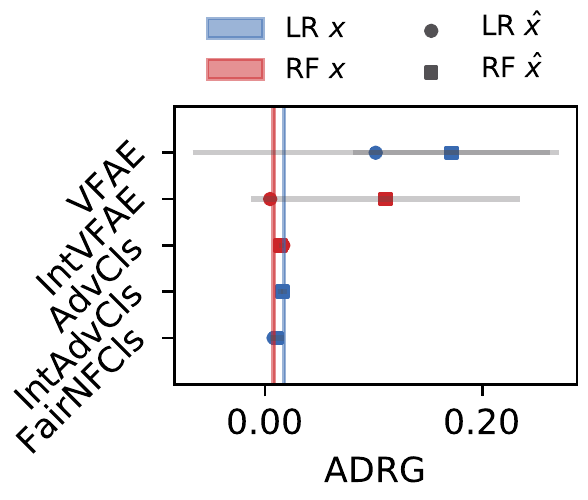}}
\subfigure[Banks Classification]{\label{fig:cls_banks_adrg}\includegraphics[width=0.32\textwidth]{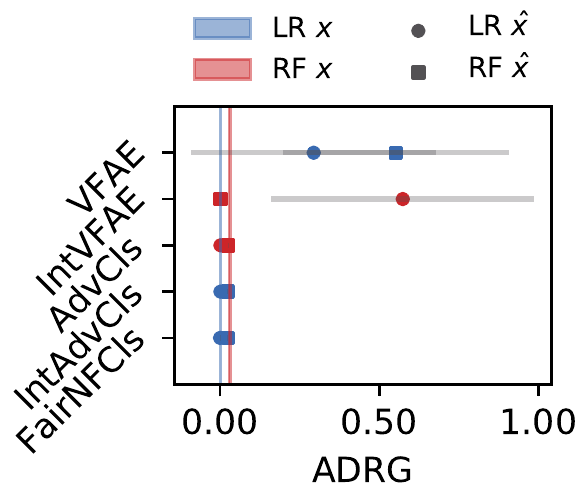}}
\caption{Performance of predicting the sensitive attribute of the external classifiers trained on the representations of the different models. The lines show the baseline on the untransformed dataset $x$, while the points show the different results on the model representations $\hat{x}$. Note that despite the ADRG metric having an asymmetric distribution, we employed a symmetric error computation.}\label{fig:classification_reps}
\end{figure*}

In fair representation learning, the ultimate objective is removing information about the sensitive attribute in the obtained representation. We investigate this matter by training off-the-shelf rankers and classifiers on the transformed data. The task is to classify the sensitive attribute, i.e. recovering information about it. We show results for this task in Figure~\ref{fig:ranking_reps} and Figure~\ref{fig:classification_reps} of the supplementary material.
The employed models are plain Linear Regression (LR) and Random Forests (RF). We evaluate the absolute difference in accuracy from the random guess (ADRG \cite{cerrato2020constraining,cerrato2020pairwise,Louizos2016TheVF}). As explained above, this metric is defined as the absolute value of the difference between the accuracy of an external classifier and the accuracy of a classifier which always predicts the majority class. Here, the rationale is that a perfectly fair representation will force a classifier to always predict the majority class as there is no other way to obtain a higher accuracy value. Therefore, lower scores are better. We also computed the AUC of the employed classifiers. 
We observe only minimal differences between the algorithms we propose in this paper and the ones already present in the literature. One issue we observed with the VFAE algorithm is that it highly degrades external classifier performance to the point that the accuracy becomes lower than the random guess. In a binary classification setting, one could invert the decisions undertaken by the model and obtain a higher performance than random guessing: therefore, this model displays fairly high ADRG.

As an aside, gains in representation invariance with respect to the sensitive attribute are very small when considering AUC. To the best of our knowledge, this is the first time that AUC has been considered for this kind of evaluation, with previous literature focusing mostly on accuracy-based measures such as ADRG (see e.g. \cite{Louizos2016TheVF,zemel2013learning,cerrato2020constraining}). We conclude that employing AUC should be strongly considered in the future when evaluating fair representation learning algorithms.

\subsection{Accuracy-Discrimination tradeoffs}

\begin{figure*}[!ht]
\centering
\subfigure[VFAE CompasCls]{\label{fig:compare_auc0}\includegraphics[width=0.32\textwidth]{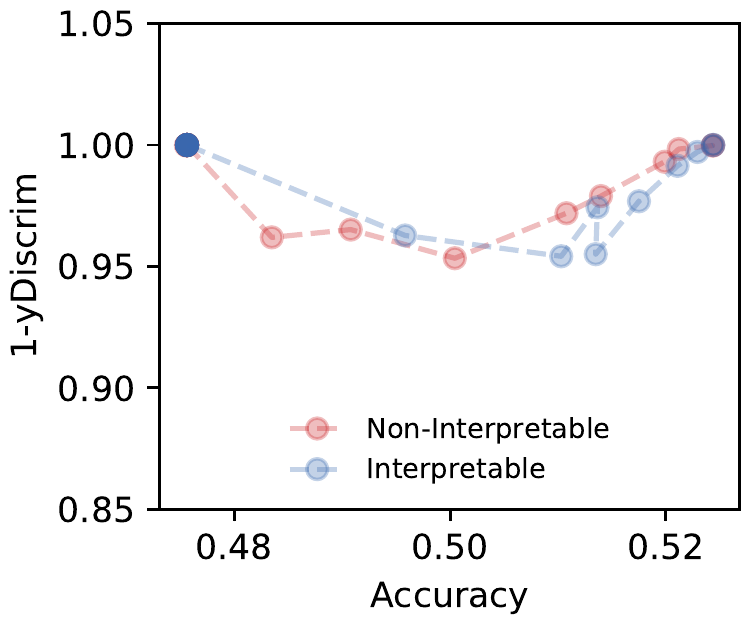}}
\subfigure[VFAE Adult]{\label{fig:compare_auc1}\includegraphics[width=0.32\textwidth]{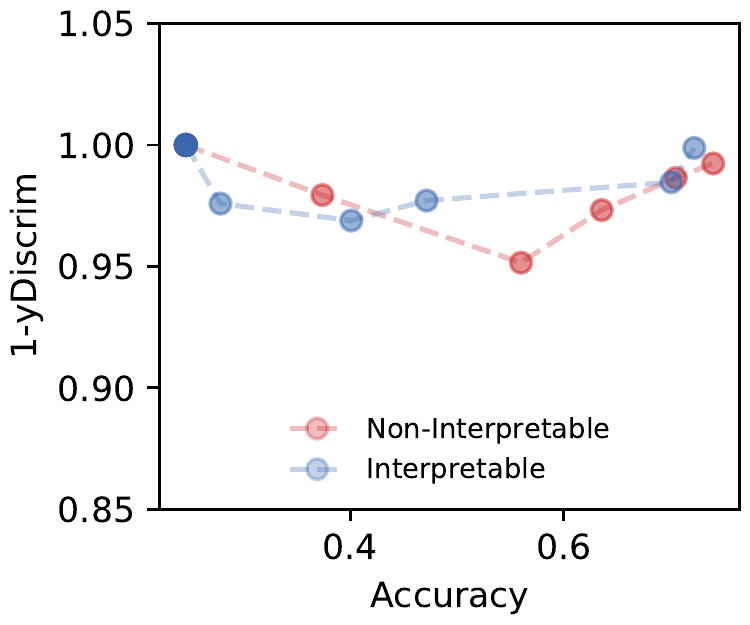}}
\subfigure[VFAE BanksCls]{\label{fig:compare_auc2}\includegraphics[width=0.32\textwidth]{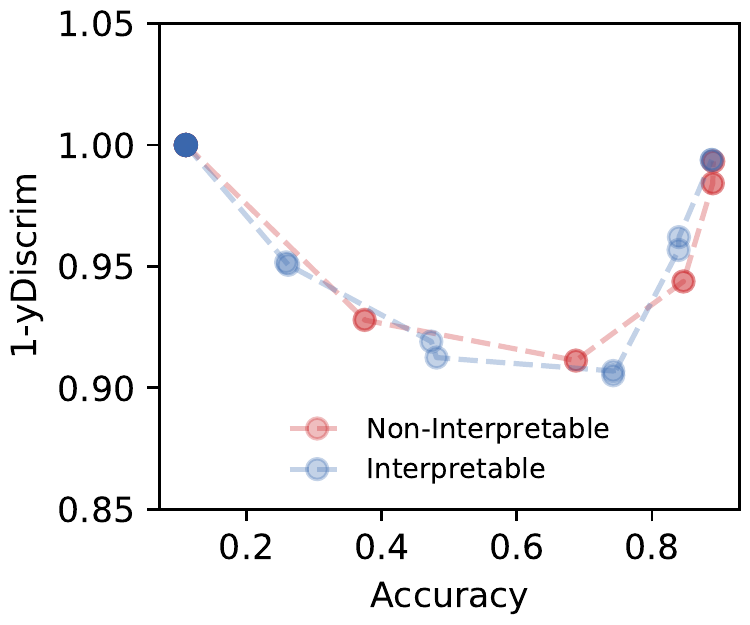}}
\subfigure[AdvCls CompasCls]{\label{fig:compare_auc3}\includegraphics[width=0.32\textwidth]{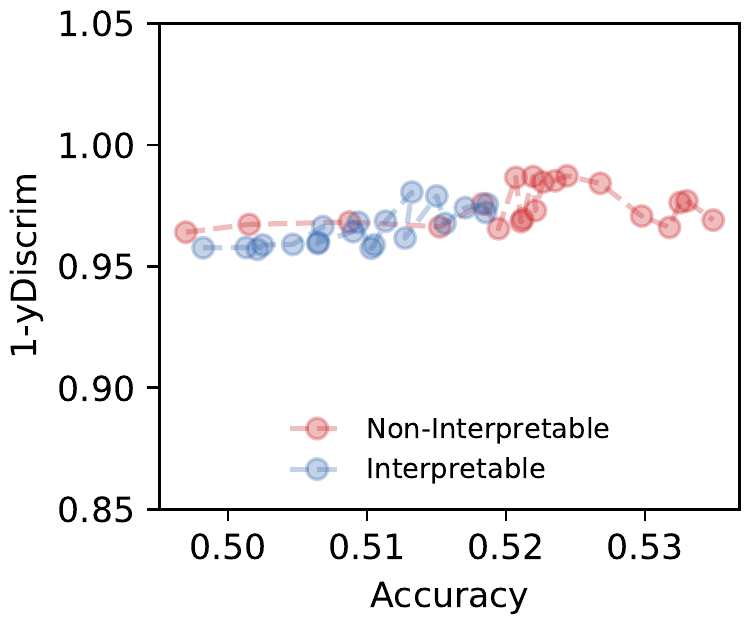}}
\subfigure[AdvCls Adult]{\label{fig:compare_auc4}\includegraphics[width=0.32\textwidth]{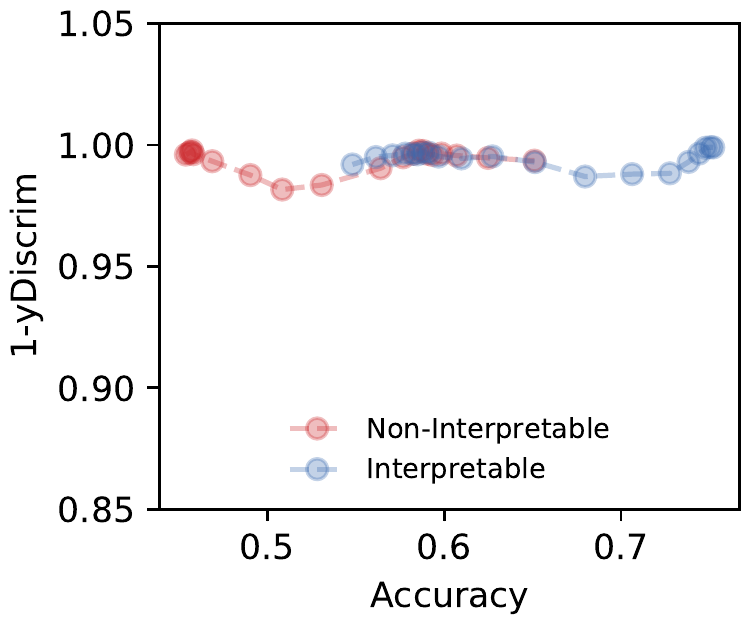}}
\subfigure[AdvCls BanksCls]{\label{fig:compare_auc5}\includegraphics[width=0.32\textwidth]{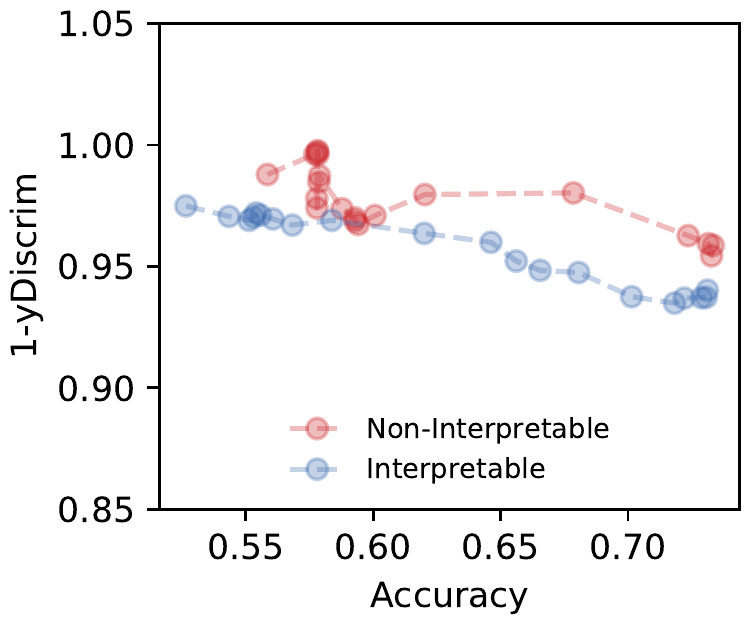}}
\caption{Accuracy/discrimination tradeoffs for our classification models. We considered 20 different thresholds in the interval [0.05, 1). Higher accuracy is better; lower discrimination is better.}\label{fig:compare_auc}
\end{figure*}

We report in Figure~\ref{fig:compare_auc} the accuracy/discrimination tradeoffs for our classification models. This analysis was performed by taking 20 different thresholds in the interval $[0.05, 1)$ and computing accuracy and discrimination metrics. The highest discrimination value (lower is better, see Section~\ref{sec:AUDC} and~\ref{sec:app_AUDC}) we find is $0.06$. 

\end{document}